\PassOptionsToPackage{numbers,sort&compress}{natbib}  
\documentclass[preprint,12pt]{elsarticle}
\usepackage[margin=2.5cm]{geometry}
\usepackage{enumitem}
\newlist{inlinelist}{enumerate*}{1}
\setlist*[inlinelist,1]{%
  label=(\roman*),
}
\usepackage{dirtytalk}
\usepackage{soul}
\usepackage{hhline}
\usepackage{subfig}
\usepackage{multirow}
\usepackage{xcolor}
\usepackage{graphicx}
\usepackage{upgreek}
\usepackage{amssymb}
\usepackage{amsfonts,amsthm,bm,amsmath} 
\usepackage{appendix}
\usepackage{times}
\usepackage{caption}
\usepackage{subfig}
\newcommand{\psubref}[1]{\protect\subref{#1}}
\usepackage{multirow}

\usepackage[colorlinks]{hyperref} 
\hypersetup{ 
    colorlinks=true,       
    linkcolor=red,          
    citecolor=blue,        
    filecolor=magenta,      
    urlcolor=cyan           
}

\newcommand{\fref}[1]{Fig.~\ref{#1}}

\newcommand{\sref}[1]{Section~\ref{#1}}
\newcommand{\tref}[1]{Table~\ref{#1}}
\setcitestyle{square}

\journal{Computers \& Structures}
\begin{document}

\begin{frontmatter}

\title{Exploring the structure-property relations of thin-walled, 2D extruded lattices using neural networks}
\author[]{Junyan He$^1$}
\author[]{Shashank Kushwaha$^1$}
\author[]{Diab Abueidda$^2$}
\author[]{Iwona Jasiuk$^1$\corref{mycorrespondingauthor}}
\address{$^1$ Department of Mechanical Science and Engineering, University of Illinois at Urbana-Champaign, Champaign, IL, USA \\
$^2$ National Center for Supercomputing Applications, University of Illinois at Urbana-Champaign, Champaign, IL, USA}
\cortext[mycorrespondingauthor]{Corresponding author}
\ead{ijasiuk@illinois.edu}
\begin{abstract}

This paper investigates the structure-property relations of thin-walled lattices under dynamic longitudinal compression, characterized by their cross-sections and heights. These relations elucidate the interactions of different geometric features of a design on mechanical response, including energy absorption. We proposed a combinatorial, key-based design system to generate different lattice designs and used the finite element method to simulate their response with the Johnson-Cook material model. Using an autoencoder, we encoded the cross-sectional images of the lattices into latent design feature vectors, which were supplied to the neural network model to generate predictions. The trained models can accurately predict lattice energy absorption curves in the key-based design system and can be extended to new designs outside of the system via transfer learning.

\section*{\bf{Graphical abstract}}
{\centering
\includegraphics[width=0.82\textwidth]{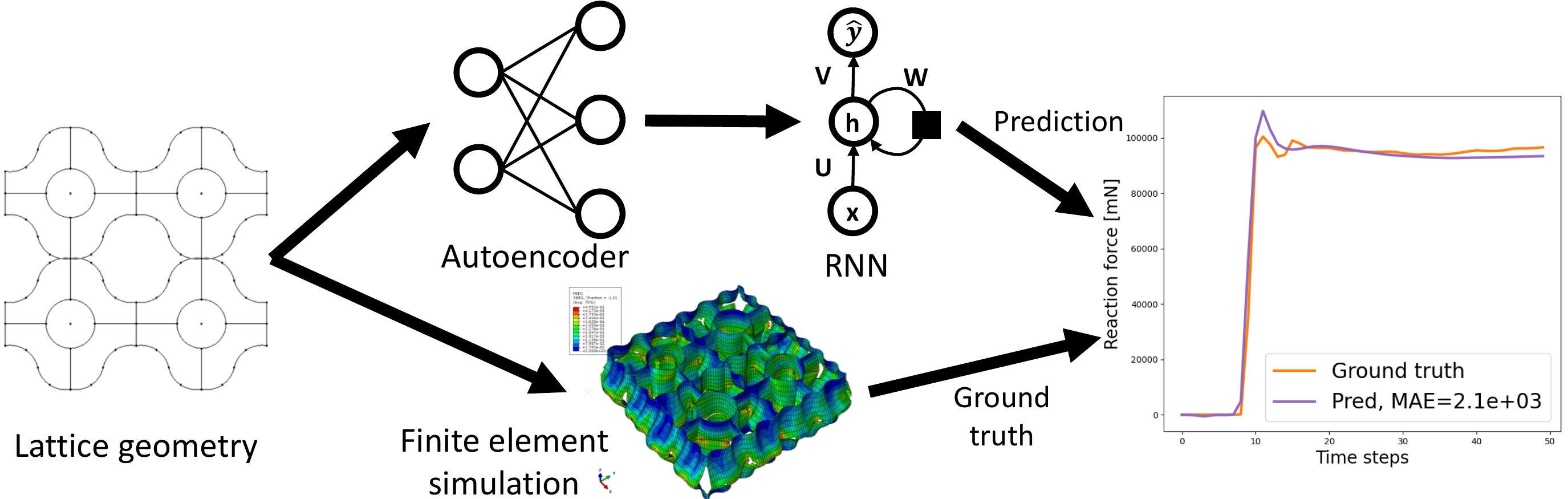}
\par
}
\end{abstract}

\begin{keyword}
Thin-walled lattices \sep Structure-property relations \sep Johnson-Cook model \sep Neural networks
\end{keyword}

\end{frontmatter}

\section{Introduction}
\label{sec:intro}
Lattice-filled sandwich panels see increasingly widespread use as energy-absorbing structures in different engineering applications such as energy absorbers \citep{shin2008experimental,xie2014impact}, sacrificial cladding \citep{van2014blast,codina2017new}, and armor plates \citep{qi2013blast,gama2001aluminum}. Previous studies have established that the core design plays a significant role in the energy absorption capabilities of the sandwich panel \citep{tarlochan2007composite}. The honeycomb \citep{tarlochan2021sandwich} lattice structure, along with many of its variants, such as the square honeycomb \citep{xue2006crush}, bio-inspired honeycombs \citep{ha2019energy}, and hierarchical honeycombs \citep{qiao2016plane}, have been studied extensively in the literature as cores of sandwich panels. They are examples of thin-walled 2D extruded lattices, whose wall thickness is significantly smaller than their in-plane dimensions, and the unit cells can be described by their cross-sectional designs and heights. To generate optimized designs that maximize specific energy absorption, parameter optimizations have been performed on design variables such as thickness \citep{paz2014crushing}, unit cell size \citep{sun2010two}, wave periodicity, and amplitude (for some bio-inspired honeycombs) \citep{ha2019energy}.

The aforementioned sizing optimization efforts and the construction of response surfaces for the cores \citep{sun2010two,panda2018experimental} provide insights into the structure-property relations relating the lattice structure to key performance metrics like stress-strain curve and energy absorption during deformation for a limited set of honeycomb-like lattice core designs. It is of great research and industrial interests to further extend such structure-property relations to a diverse set of lattice core designs with different cross-sectional geometries, and in particular, to elucidate how different geometric features such as the addition of wavy unit-cell walls, hierarchical designs, auxetic designs, and fractal designs combine and interact to affect the specific energy absorption of the lattice core design. The exploration of structure-property relations inherently involves surveying many different lattice core designs. Most current thin-walled lattice core designs in the literature were generated based on engineering intuition, experiments, and/or bio-inspiration \citep{CHRISTENSEN200093,gao2018topological,abueidda2020topology}. Other studies leveraged optimization-based methods like the hybrid cellular automata \citep{duddeck2016topology,zeng2017improved}, ground structure approach \citep{guo2021topology}, and the ant colony optimization method \citep{sharafi2014shape} to systematically generate new, optimized cross sections for lattices under longitudinal compression. Nonetheless, there is a lack of a systematic compilation of the mechanical response and energy absorption characteristics for these new designs and an attempt to reveal potential structure-property relations for various lattice cross-sectional designs.

Our current work aims to develop a systematic framework to generate distinct lattice cross-section designs for longitudinal compression. To explore the structure-property relations, the considered structures do not need to have optimized energy absorption characteristics. Thus, we chose to generate new lattice cross-section designs via a simple combinatorial framework. Combinatorial methods have been applied to generate trussed-based lattices \citep{verma2020combinatorial,bastek2022inverting,baykasouglu2020multi}, triply periodic minimal surface lattices \citep{wang2021hierarchical,callanan2018hierarchical}, and thin-walled structures \citep{wang2018six}, which create new designs by combining different pre-selected geometric descriptors. It is a novel attempt to leverage a combinatorial framework to generate 2D extruded lattice core designs for longitudinal compression systematically. The second objective of this work is to approximate the underlying structure-property relations via a neural network (NN) model. NN models have seen increasing use in solid mechanics, such as to predict stress-strain curve and toughness of composites \citep{yang2020prediction,gu2018novo,chen2019machine,yang2019using,ABUEIDDA2019111264} and to predict properties of lattices and rapidly evaluate performance \citep{bastek2022inverting,laban2020experimental, 10.1115/1.4045040,garland2020deep,hassanin2020controlling}. However, NN models have not been widely applied to approximate the structure-property relations of thin-walled lattice cores.

This paper is organized as follows: \sref{sec:methods} presents an overview of the combinatorial generation framework, numerical simulation, preprocessing of the design images, and the architecture of the NN model. \sref{sec:results} presents the results, and \sref{sec:discussion} discusses the quality of the image preprocessing and NN predictions. \sref{sec:conc} summarizes the outcomes and highlights possible future works.

\section{Methods}
\label{sec:methods}
\subsection{A framework to create lattices using geometric descriptors}
\label{sec:geometric}
Diverse design space of lattice cross-sections is needed such that the subsequent NN can extract trends from various lattice architectures. Drawing inspiration from the truss descriptors used to construct truss-based lattices in Bastek et al. \citep{bastek2022inverting} and Zok et al. \citep{zok2016periodic}, we define a set of geometric descriptors for generating lattice cross-sections. In this work, we focus on 2D extruded lattices with constant cross-sections. Such lattices can be uniquely defined by their cross-sections and heights. We further require that the unit cell of the periodic lattice fits within a square bounding box. In this case, a 2D tessellation can be generated by repeating the unit cell along the X and Y directions. With this simplification, one can formulate three types of geometric descriptors:
\begin{itemize}
    \item Vertex. The vertices of the unit cell bounding box can take on different styles: it can be a vertex or the vertex can be replaced by a straight edge or an arc, see \fref{vertex_style}. 
    \begin{figure}[h!]
        \newcommand\x{0.15}
        \centering
         \subfloat[]{
             \includegraphics[width=\x\textwidth]{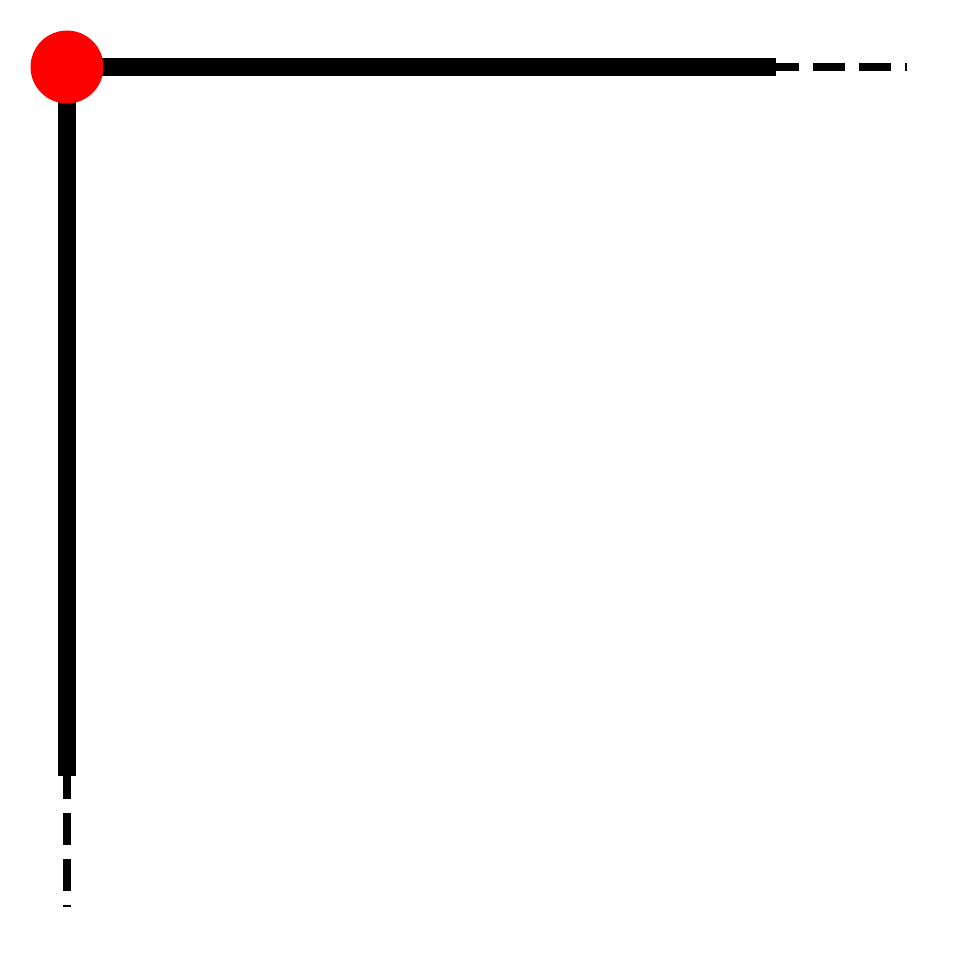}
             \label{fig:vs1}
         }
         \subfloat[]{
             \includegraphics[width=\x\textwidth]{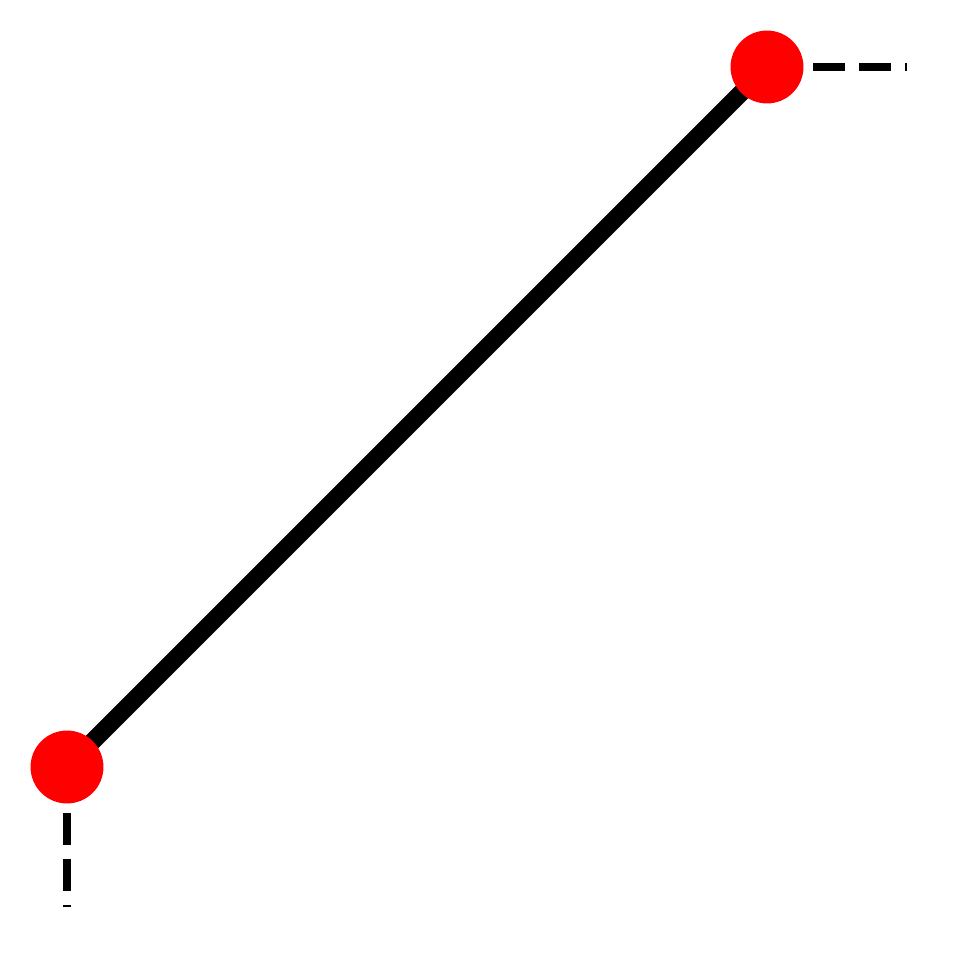}
             \label{fig:vs2}
         }
         \subfloat[]{
             \includegraphics[width=\x\textwidth]{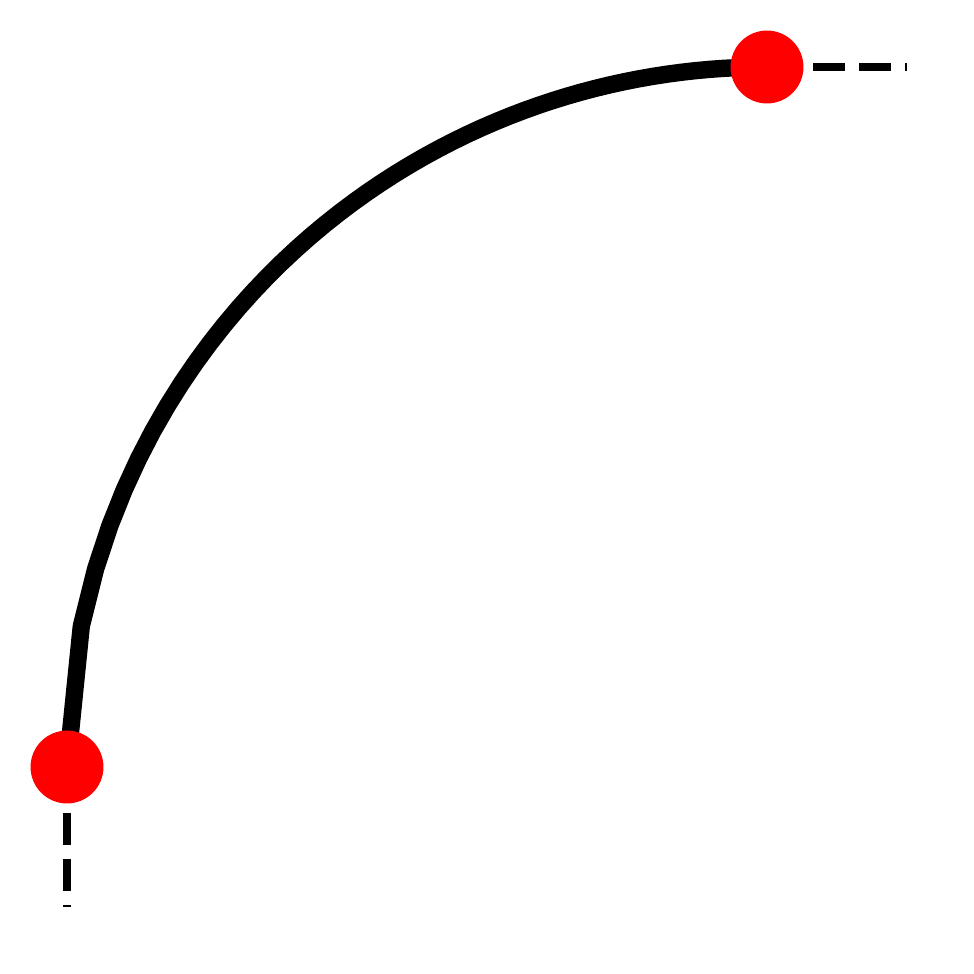}
             \label{fig:vs3}
         }
        \caption{ \psubref{fig:vs1} The vertex of the unit cell remains as-is. \psubref{fig:vs2} The vertex is replaced by a straight edge. \psubref{fig:vs3} The vertex is replaced by an arc.}
        \label{vertex_style}
    \end{figure}

    \item Edge segment. Each edge of the unit cell bounding box can be divided into different edge segments, and each can take on a different style: it can be a straight edge or two arcs (inspired by the bio-inspired honeycomb studied in the work of Ha et al. \citep{ha2019energy}), see \fref{edge_style}. 
    \begin{figure}[h!]
        \newcommand\x{0.25}
        \centering
         \subfloat[]{
             \includegraphics[width=\x\textwidth]{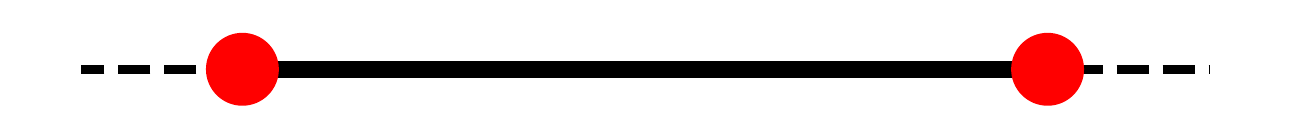}
             \label{fig:es1}
         }
         \subfloat[]{
             \includegraphics[width=\x\textwidth]{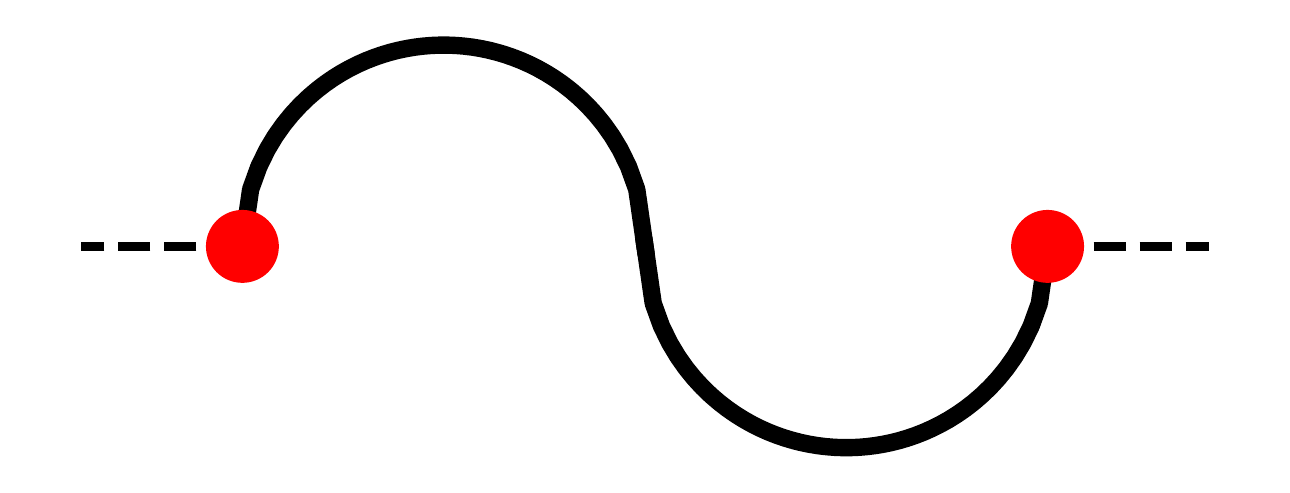}
             \label{fig:es2}
         }
        \caption{ \psubref{fig:es1} The edge segment of the unit cell remains as-is. \psubref{fig:es2} The edge segment is replaced by two arcs.}
        \label{edge_style}
    \end{figure}

    \item Interior support. Supports can be added to the interior of the unit cell: it can be no support, a $+$-shape, or an $X$-shape support. A circle can be added to the center of the unit cell, see \fref{int_style}. 
    \begin{figure}[h!]
        \newcommand\x{0.15}
        \centering
         \subfloat[]{
             \includegraphics[width=\x\textwidth]{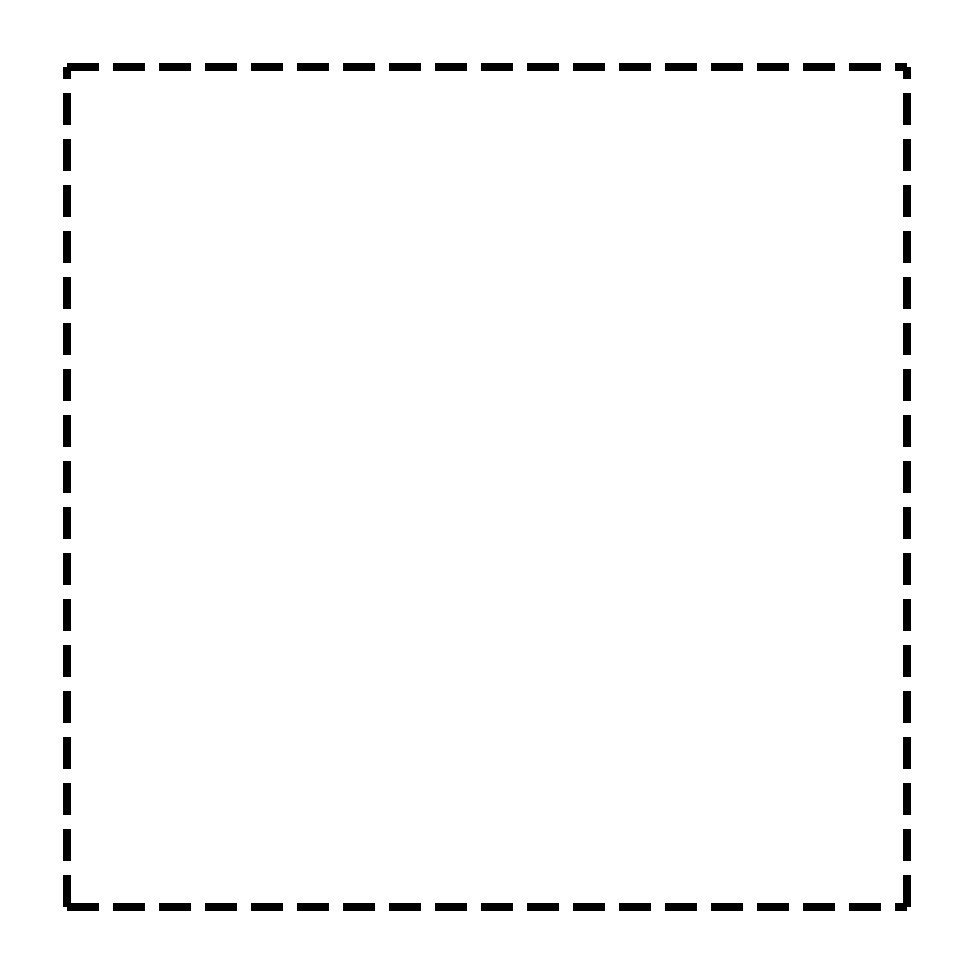}
             \label{fig:is1}
         }
         \subfloat[]{
             \includegraphics[width=\x\textwidth]{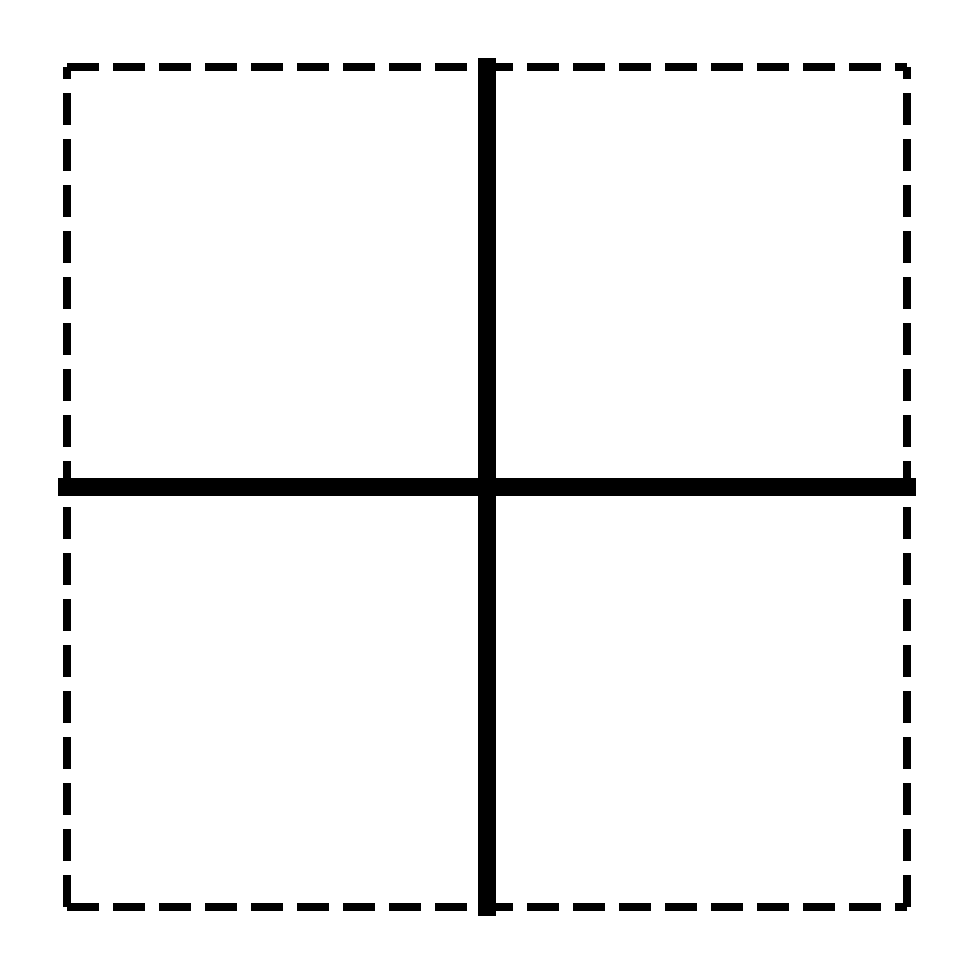}
             \label{fig:is2}
         }
         \subfloat[]{
             \includegraphics[width=\x\textwidth]{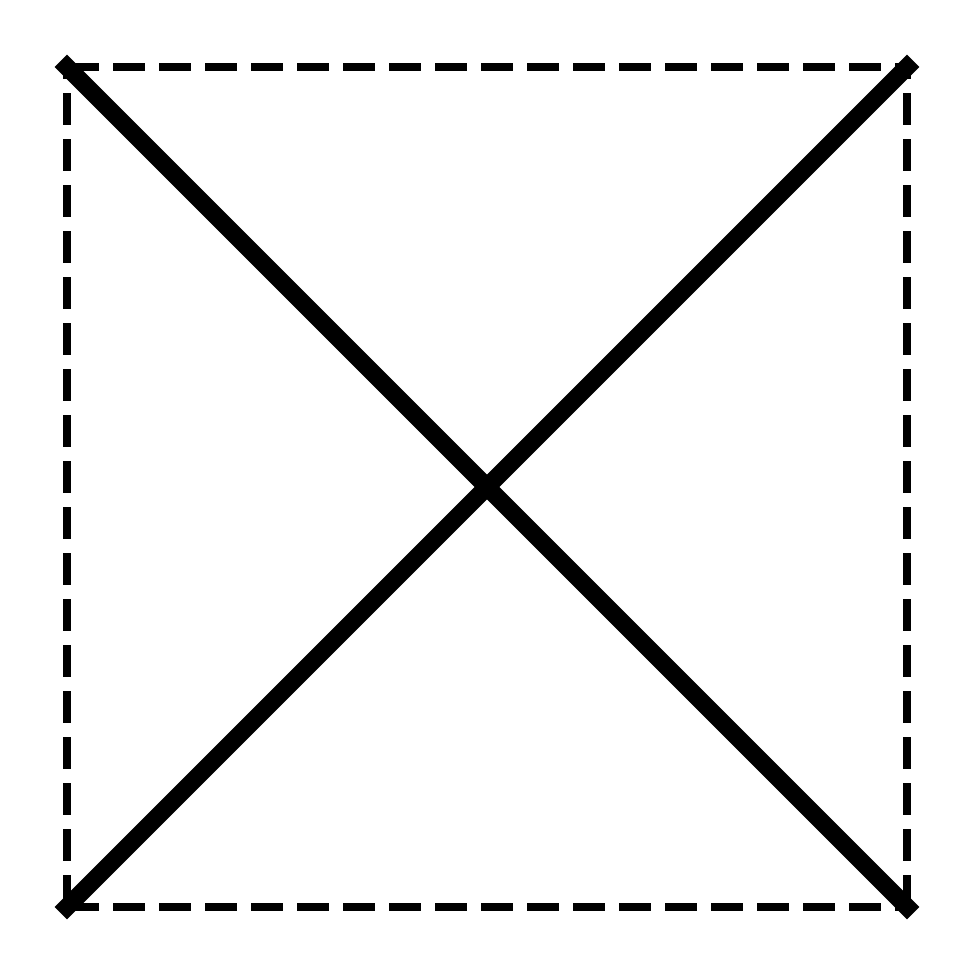}
             \label{fig:is3}
         }
         \subfloat[]{
             \includegraphics[width=\x\textwidth]{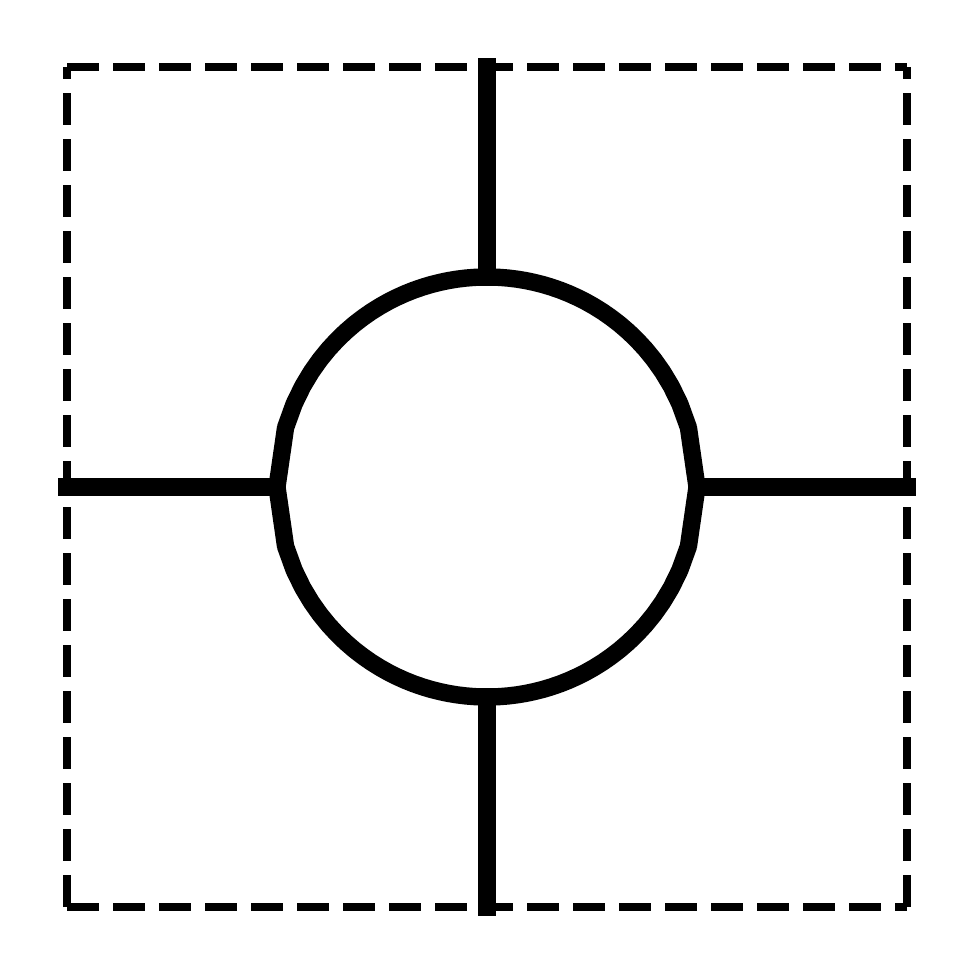}
             \label{fig:is4}
         }
         \subfloat[]{
             \includegraphics[width=\x\textwidth]{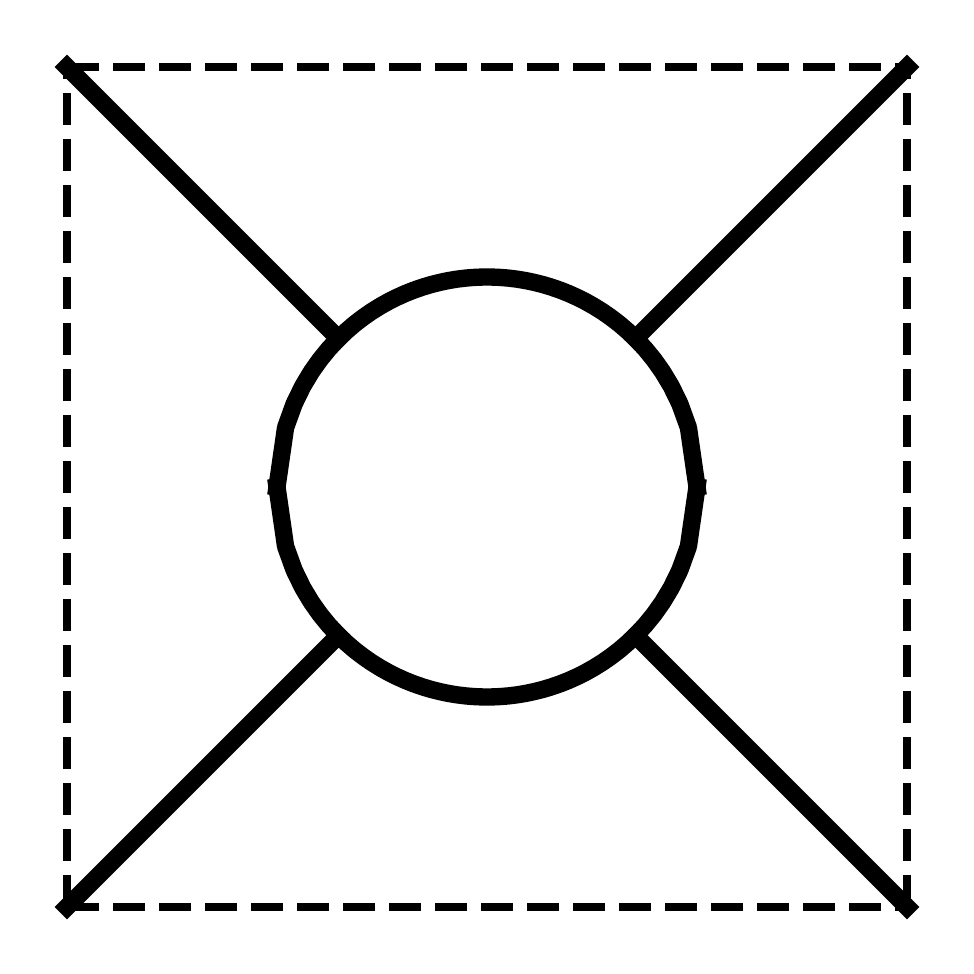}
             \label{fig:is5}
         }
        \caption{ \psubref{fig:is1} No interior support is added. \psubref{fig:is2} A $+$-shape support is added. \psubref{fig:is3} A $X$-shape support is added. \psubref{fig:is4} A $+$-shape support is added along with a circle. \psubref{fig:is5} A $X$-shape support is added along with a circle.}
        \label{int_style}
    \end{figure}
\end{itemize}
It is important to point out that the lattice geometric descriptors are not unique; other choices are possible to generate a different design space. The descriptors and their sub-options chosen here aim to generate lattices that bear similar design cues to some traditional or bio-inspired lattices seen in the literature \citep{ha2019energy,yang2018out,san2020review}.

\begin{figure}[h!]
    \newcommand\x{0.3}
    \centering
    \begin{tabular}{ c  c  c  }
    \begin{minipage}[c]{\x\textwidth}
       \centering 
        \subfloat[Design 00231121]{\includegraphics[width=\textwidth]{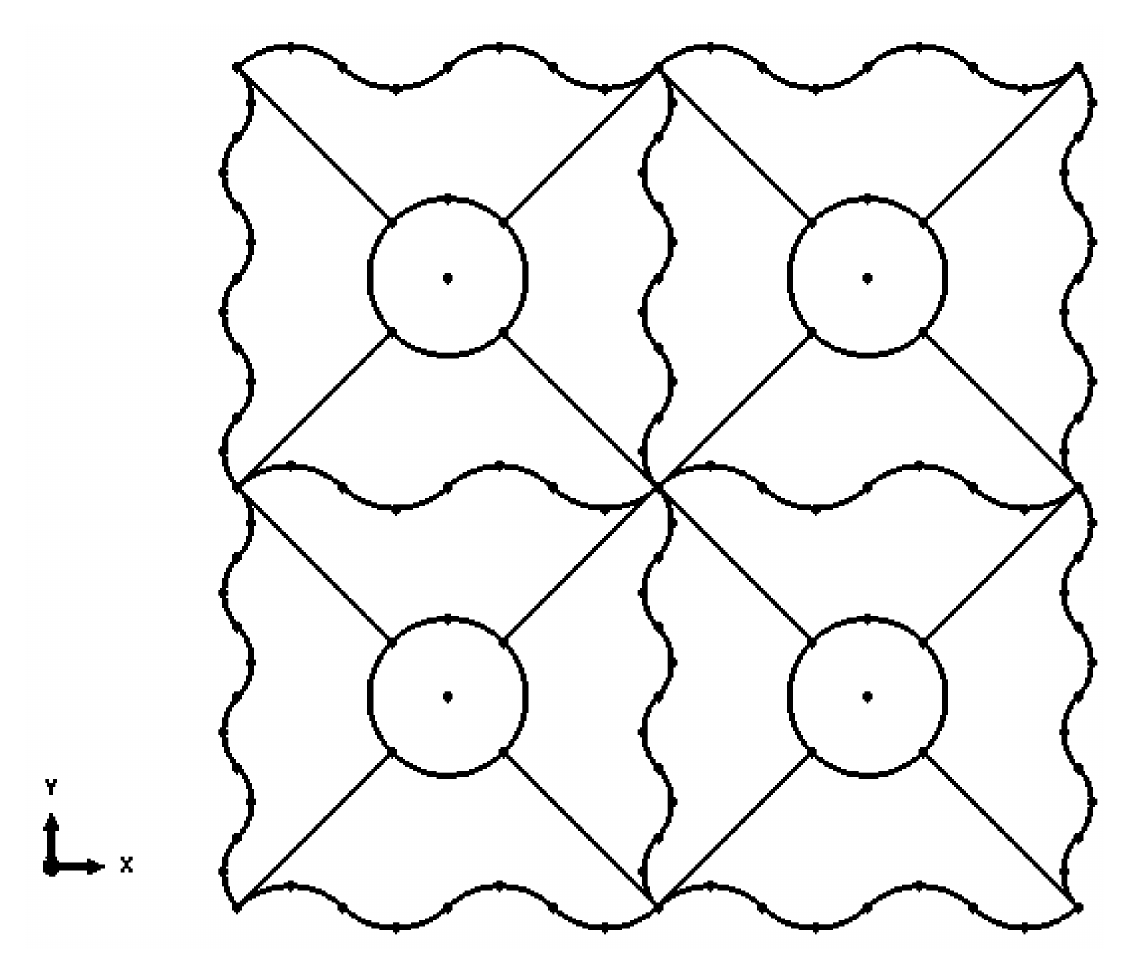}
        \label{fig:d1}}
    \end{minipage}
    &
    \begin{minipage}[c]{\x\textwidth}
       \centering 
        \subfloat[Design 10331011]{\includegraphics[width=\textwidth]{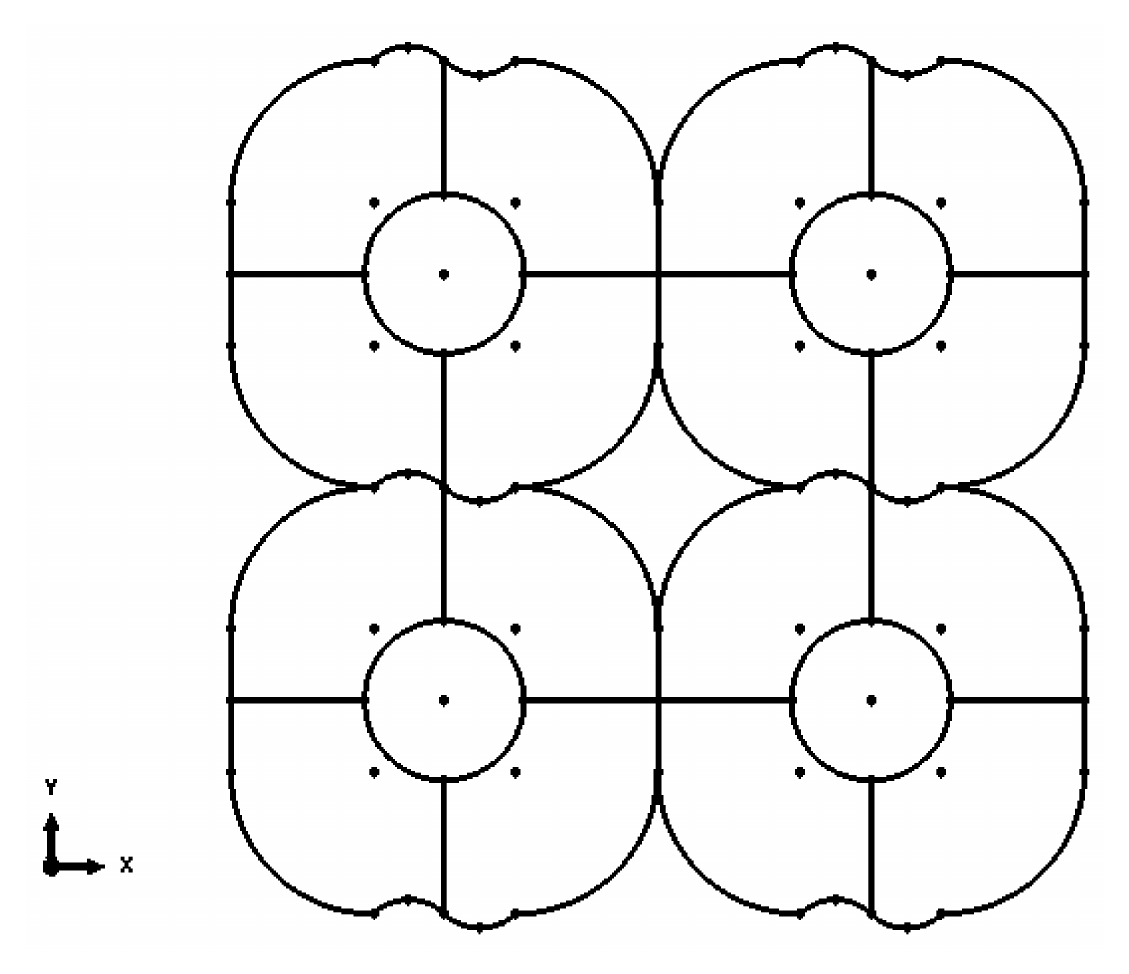}
        \label{fig:d2}}
    \end{minipage}
    &
    \begin{minipage}[c]{\x\textwidth}
       \centering 
        \subfloat[Design 11220020]{\includegraphics[width=\textwidth]{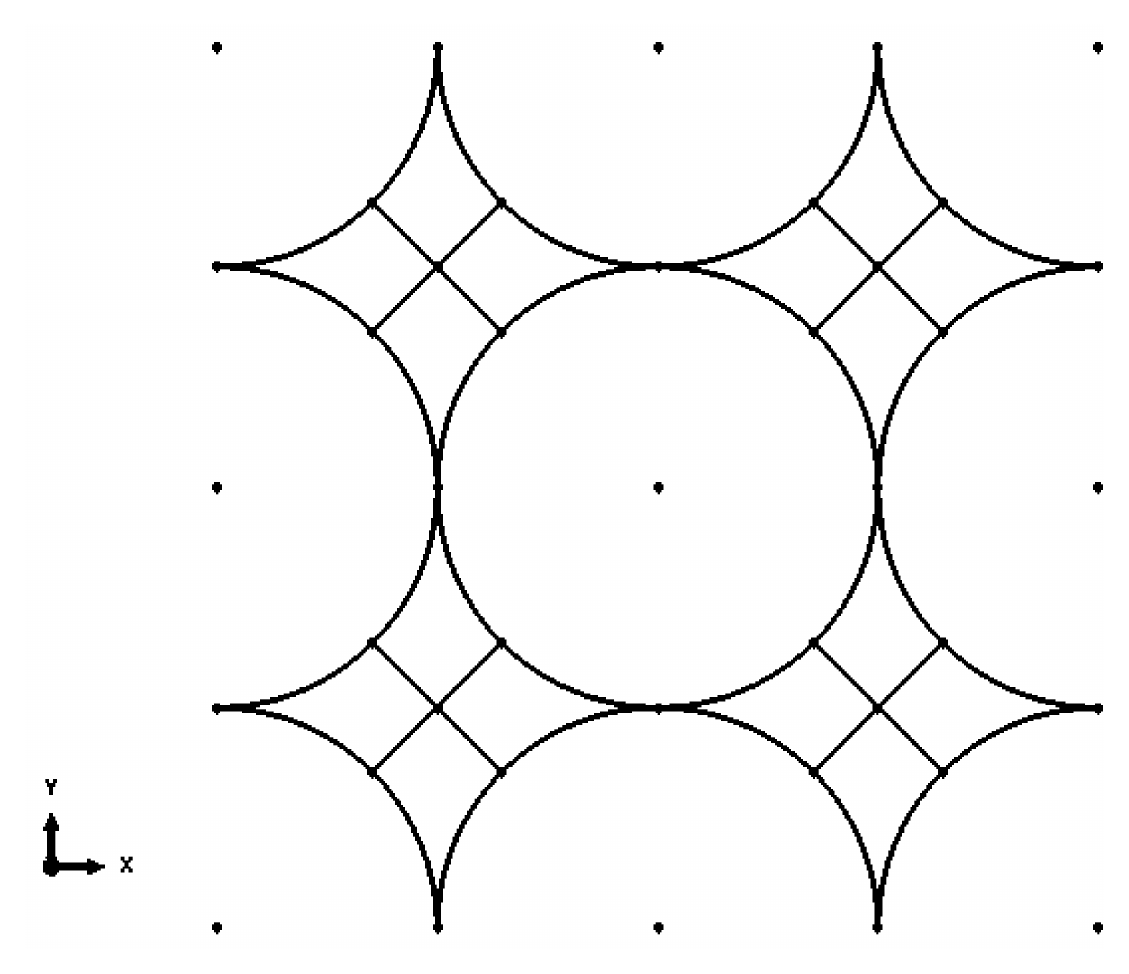}
        \label{fig:d3}}
    \end{minipage}\\

    \begin{minipage}[c]{\x\textwidth}
       \centering 
        \subfloat[Design 11331011]{\includegraphics[width=\textwidth]{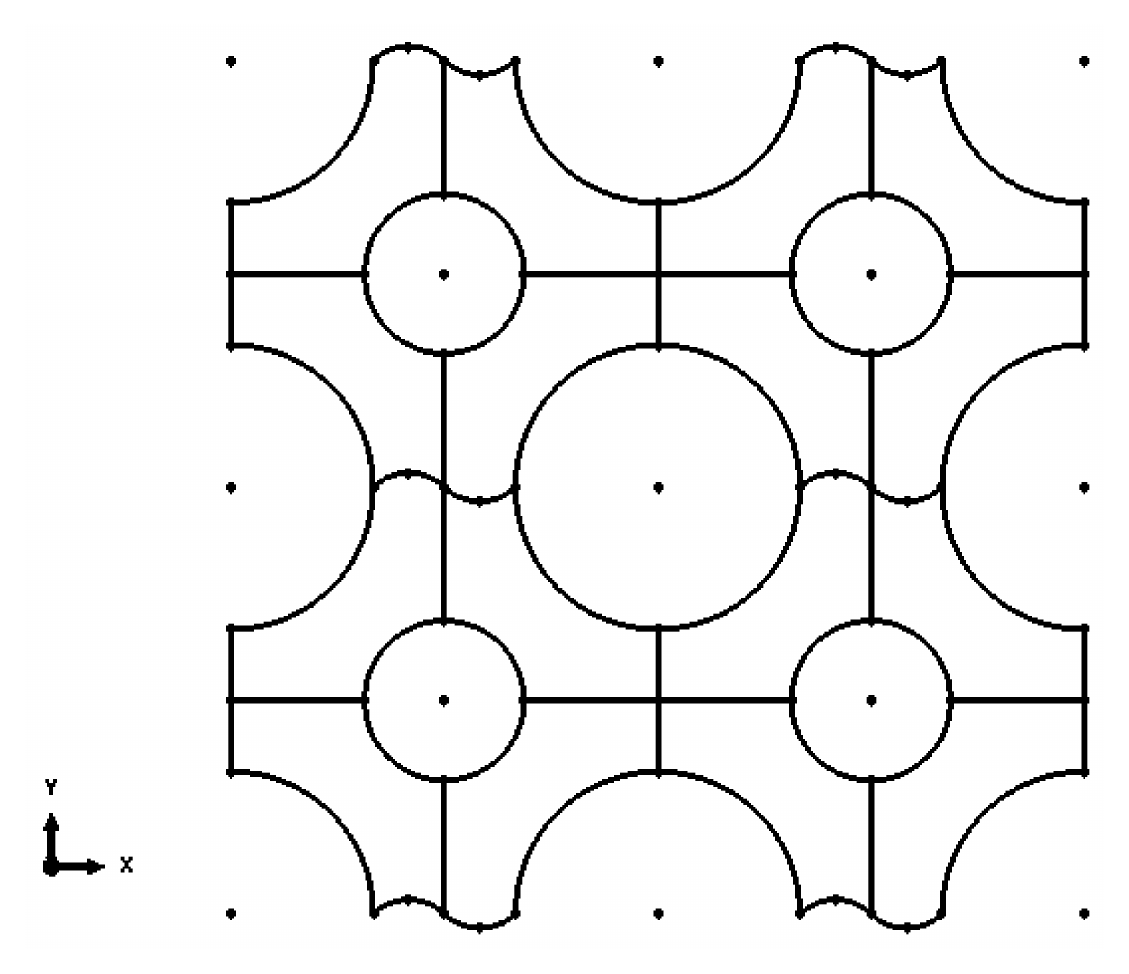}
        \label{fig:d4}}
    \end{minipage}
    &
    \begin{minipage}[c]{\x\textwidth}
       \centering 
        \subfloat[Design 20220000]{\includegraphics[width=\textwidth]{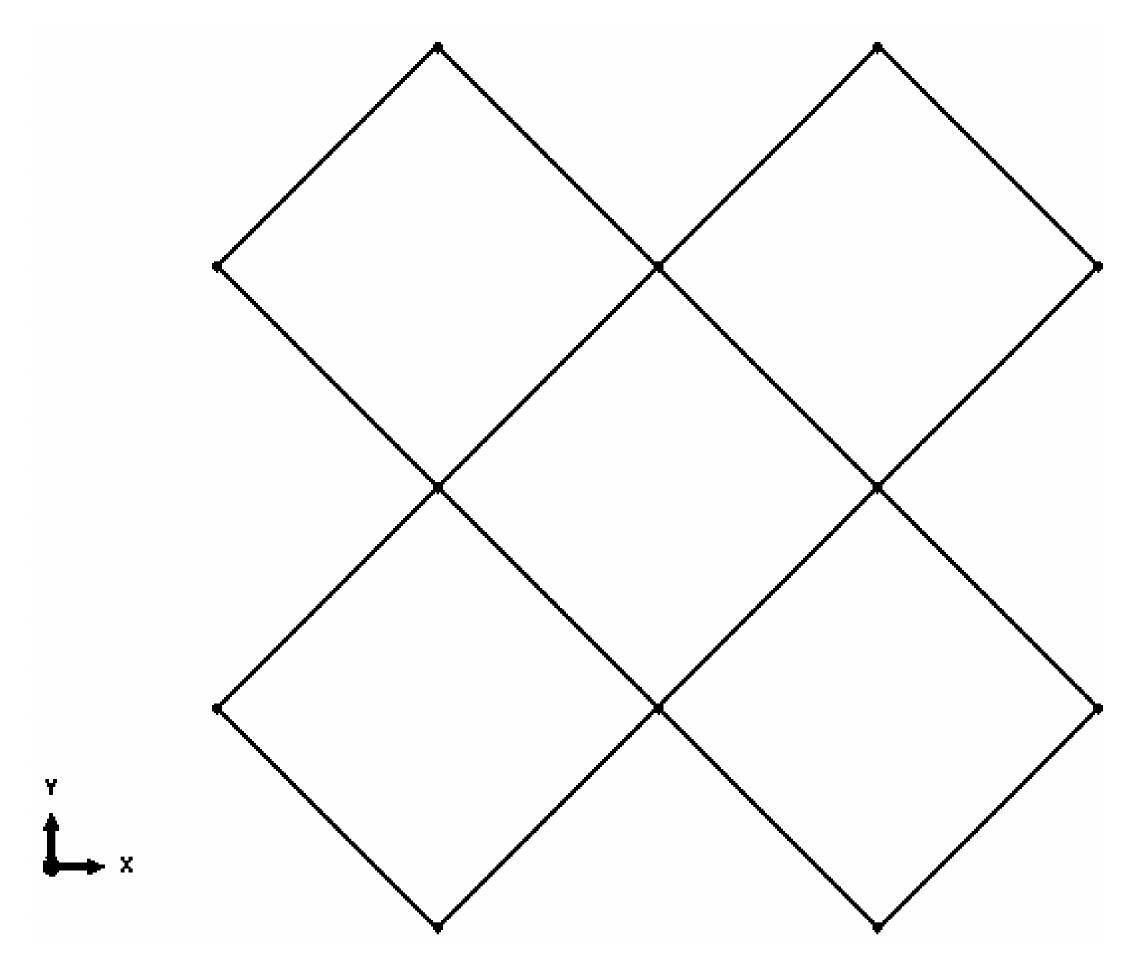}
        \label{fig:d5}}
    \end{minipage}
    &
    \begin{minipage}[c]{\x\textwidth}
       \centering 
        \subfloat[Design 21221110]{\includegraphics[width=\textwidth]{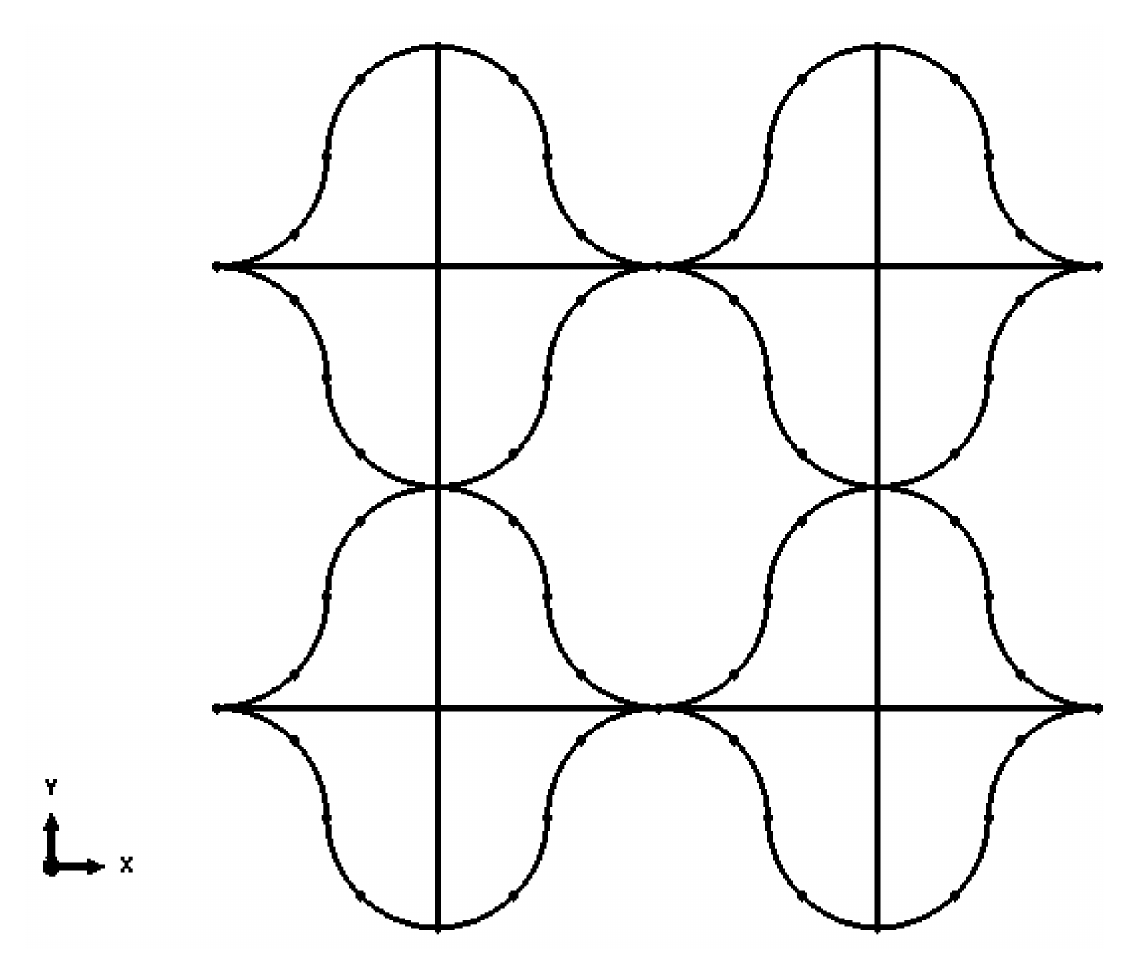}
        \label{fig:d6}}
    \end{minipage}
    \end{tabular}
    \caption{A survey of some lattice designs generated by the key-based system.}
    \label{survey}
\end{figure}

A unique lattice design can then be generated in a combinatorial manner by forming different combinations of the geometric descriptors. Some geometric descriptors also allow for sub-design options. In the straight edge vertex case (\fref{fig:vs2}), the edge can take on either edge style depicted in \fref{edge_style}. In the arc vertex case (\fref{fig:vs3}), the arc can point inward or outward to the unit cell. Vertical and horizontal edges of the unit cell bounding box can be divided into a different number of edge segments. Still, we limit to a minimum of 2 and a maximum of 4 divisions for all edges. The lattice design created by a combination of the design options and their sub-options can be conveniently denoted by an 8-digit key. Each digit represents a choice on each option/sub-option. A list of all possible options is presented in \tref{design_tab}. This key-based design system contains 660 unique lattice design keys. Besides the discrete design variables, the framework also features a continuous design variable, namely the thickness of the lattice walls. In the analysis, we fixed the number of unit cells to be 4 in the cross-section, forming a 2-by-2 arrangement. Note that different numbers of unit cells and their arrangements can be used to construct the lattice structure and will affect the mechanical response of the lattice. Studying the effects of the number of unit cells and unit cell arrangement is a subject of our future works. A survey of some periodic lattices generated from the system is presented in \fref{survey}.

\begin{table}[h]
    \caption{Explanation of the 8-digit design key system}
    \small
    \centering
    \begin{tabular}{cccccccccc}
    \hline
    Digit & \vline & 1 & 2 & 3 & 4 & 5 & 6 & 7 & 8 \\
    
    \hline
    Option & \vline & \begin{tabular}{@{}c@{}}Vertex\end{tabular} & \begin{tabular}{@{}c@{}}Arc\\sub-opt\end{tabular} & \begin{tabular}{@{}c@{}}\# segments\\horizontal\end{tabular} & \begin{tabular}{@{}c@{}}\# segments\\vertical\end{tabular} &
    \begin{tabular}{@{}c@{}}Edge\\sub-opt\\horizontal\end{tabular} & \begin{tabular}{@{}c@{}}Edge\\sub-opt\\vertical\end{tabular} & Interior & \begin{tabular}{@{}c@{}}Interior\\sub-opt\end{tabular} \\ 
    
    \hline
    \begin{tabular}{@{}c@{}}Possible\\values\end{tabular} & \vline & 
    \begin{tabular}{@{}c@{}}0 (\ref{fig:vs1})\\1 (\ref{fig:vs2})\\2 (\ref{fig:vs3})\end{tabular} & 
    \begin{tabular}{@{}c@{}}0 (in)\\1 (out)\end{tabular} &
    \begin{tabular}{@{}c@{}}2\\3\\4\end{tabular} & 
    \begin{tabular}{@{}c@{}}2\\3\\4\end{tabular} & 
    \begin{tabular}{@{}c@{}}0 (\ref{fig:es1})\\1 (\ref{fig:es2})\end{tabular} &
    \begin{tabular}{@{}c@{}}0 (\ref{fig:es1})\\1 (\ref{fig:es2})\end{tabular} &
    \begin{tabular}{@{}c@{}}0 (\ref{fig:is1})\\1 (\ref{fig:is2})\\2 (\ref{fig:is3})\end{tabular} & 
    \begin{tabular}{@{}c@{}}0 (\ref{fig:is4})\\1 (\ref{fig:is5})\end{tabular} \\
    
    \hline
    \end{tabular}
    \label{design_tab}
\end{table}

\subsection{Finite element simulations}
\label{sec:fe_sim}
A Python script was developed to generate cross-sectional sketches in the finite element (FE) analysis package Abaqus \citep{Abaqus2021} based on given design keys. The sketches were scaled so that the 2-by-2 lattice touches a square bounding box with a side length of 20 mm. Three-dimensional shell parts were created in Abaqus via extrusion to a fixed height of 10 mm. All lattice designs were discretized with 4-node shell elements with reduced integration (S4R) and uniform mesh sizes of 0.25 mm (in the X-Y plane) and 0.8 mm (along the height of the lattice structure) were used.

\begin{figure}[h!] 
    \centering
     \subfloat[]{
         \includegraphics[trim={0 0 18cm 0},clip,width=0.45\textwidth]{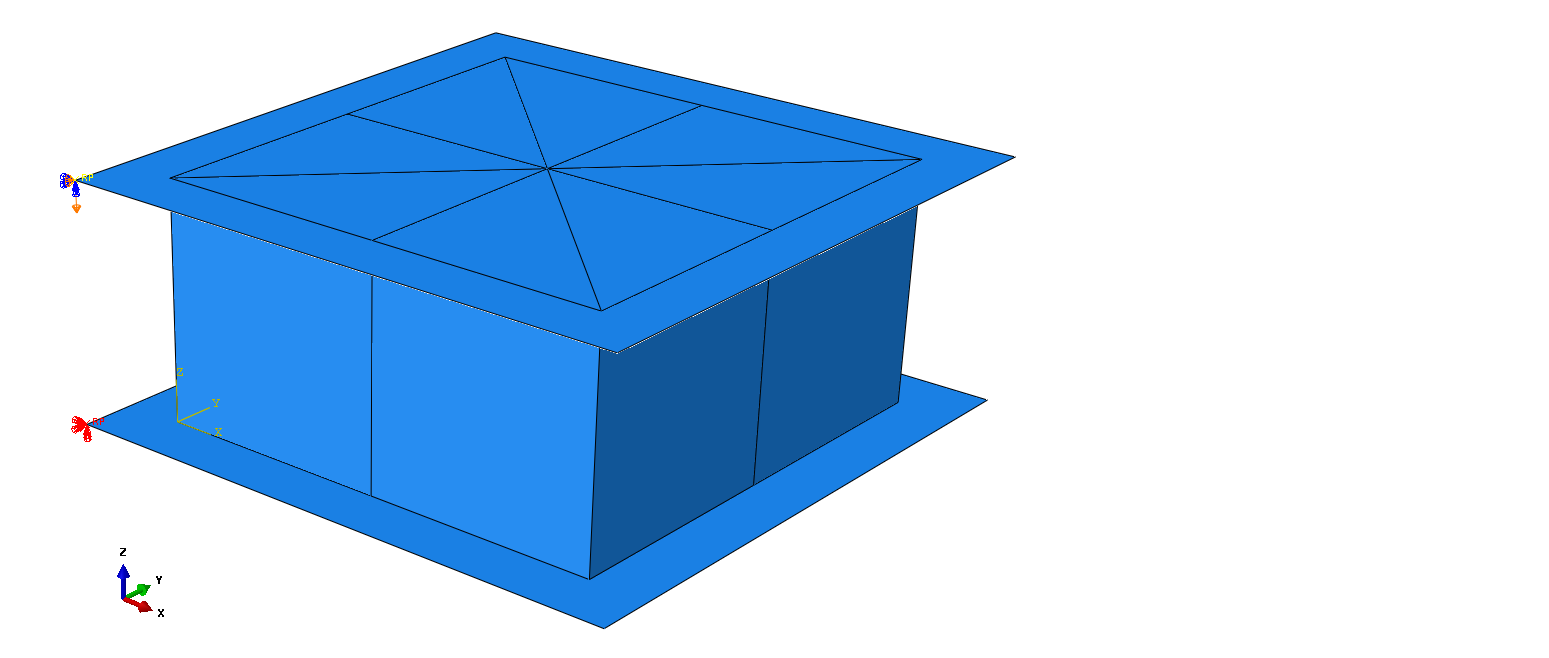}
         \label{fig:cae}
     }
     \subfloat[]{
         \includegraphics[trim={0 0 18cm 0},clip,width=0.52\textwidth]{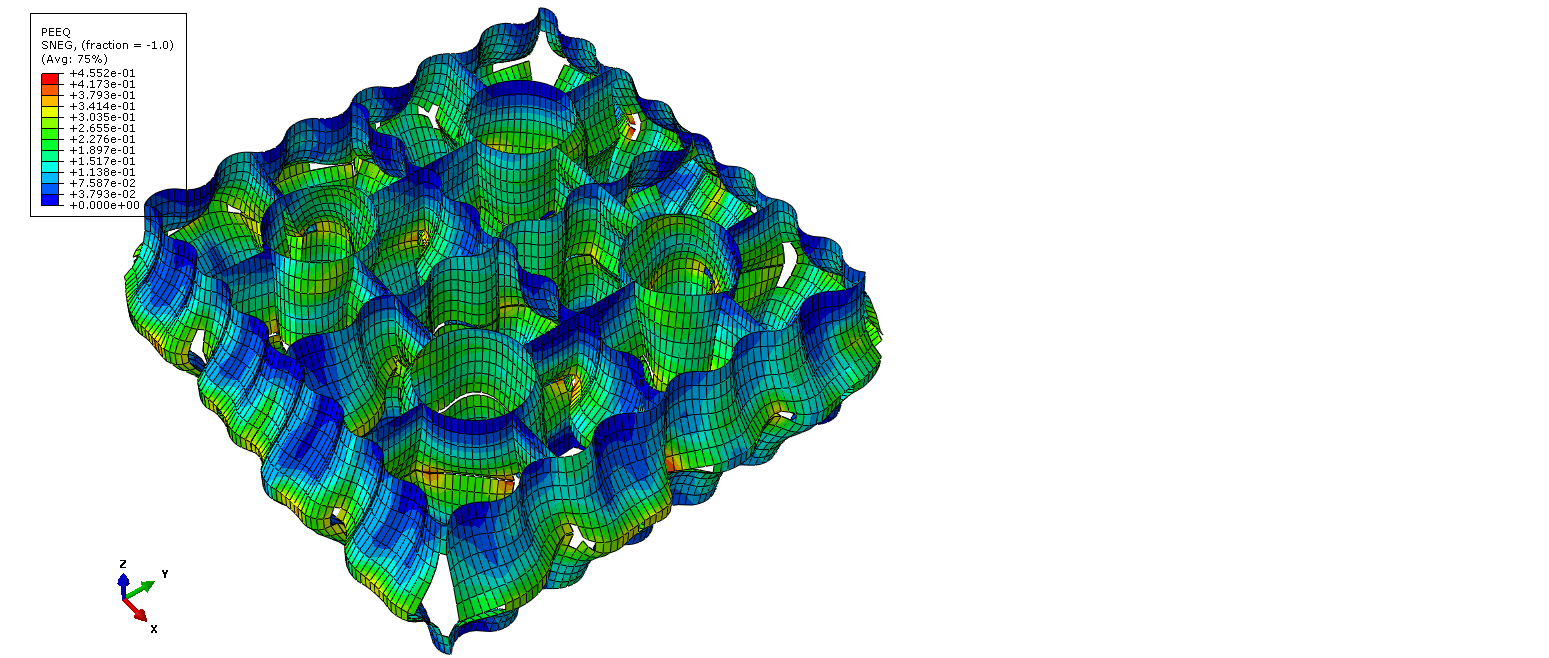}
         \label{fig:peeq}
     }
    \caption{FE model setup and results: \psubref{fig:cae} Lattice structure and two rigid plates set up for dynamic longitudinal compression. \psubref{fig:peeq} A typical deformed lattice at the end of the simulation, colored by the equivalent plastic strain.}
    \label{cae}
\end{figure}

To accurately capture the response of lattices during high strain-rate impact loading, which is common in lattice-filled sandwich panels applications, the Johnson-Cook plasticity and damage models developed by Johnson and Cook \citep{johnson1985fracture} were used in the Abaqus/Explicit dynamic simulations. The material is Ti-6Al-4V, whose properties follow from those used in the work of Wang and Shi \citep{wang2013validation} (see Table 2 therein). As a simple approximation to high strain-rate impact loading, we sandwiched the lattice between two rigid plates, and the lattices were subjected to dynamic longitudinal compression. The bottom plate, where the reaction force was measured, was held fixed, and the top plate traveled downward with a constant velocity determined by the user-defined strain rate. All sidewalls were traction-free and were free to deform. All simulations shared a constant final displacement of 2 mm, corresponding to 20\% nominal compressive strain along the Z-axis. Mass scaling was applied to the lattice to shorten simulation run time so that a minimum stable time increment of $2\times10^{-8}$ s was achieved. The reaction force at the bottom plate, displacement of the top plate, plastic dissipation, damage dissipation, and elastic strain energy of the lattice structure were outputs of the FE simulations. \fref{cae} depicts the FE model assembly and a typical deformed lattice at the end of dynamic compression.

A total of 15000 simulations were conducted on an Intel i7-11800H processor with 8 cores. Each simulation took about one to two minutes to complete. The design key, nominal strain rate, and thickness of the shell elements were sampled using a random number generator. The nominal strain rates were uniformly sampled on the log scale from the range [$10^2,10^5$] s$^{-1}$, representating strain rates during blast loading \citep{othman2016strain}. Wall thicknesses of the lattice designs were uniformly sampled from the range [0.25,0.75] mm.

\subsection{Encoding lattice cross-sectional images}
\label{sec:encoding}
As a general way to contain the geometry information in the lattice cross-section, we converted the 2D cross-sectional sketch into a 128-by-128 binary image revealing the skeleton of the cross-section. Note that the image did not include the shell thickness, so the resulting skeleton is always a single pixel wide. However, this means that the image matrices are highly sparse, and a na\"ive attempt to directly feed the image matrix to the NN as input is likely wasteful on computer memory. 

In this work, we employed a simple autoencoder to extract information in the lattice cross-sectional image. The encoder consists of two dense layers, with each having 100 neurons with the rectified linear unit as an activation function. It transforms a 128-by-128 input image into a 100-by-1 latent feature vector. The decoder consists of a single dense layer with 128$^2$ neurons and uses the sigmoid function as activation. The autoencoder was developed and tested in Keras \citep{chollet2015keras} with a TensorFlow \citep{tensorflow2015-whitepaper} backend. To train and test the autoencoder, 600 design images were randomly chosen from a total of 660 designs, and an 80-20 split was adopted for training and testing. We used the mean squared error (MSE) as the loss function and the Adam optimizer \citep{kingma2014adam} to minimize it. The autoencoder was trained for 80 epochs with a batch size of 50. To gauge the ability of the autoencoder to encode lattice structures never seen in a training set, an additional testing set consisting only of the 60 lattice designs not included in the training data set, was formed. 

\subsection{Neural network for sequence prediction}
\label{sec:NN}

\subsubsection{Input data, data augmentation, and loss function}
\label{sec:data_loss}
A sampling of the input space was described in \sref{sec:fe_sim}. The output arrays were extracted from Abaqus and were downsampled from 100 time steps down to 50-time steps to reduce size of the training data set. 
Two groups of inputs were generated for the NN model.
The first group contains information on the lattice cross-sections in the form of 100-by-1 encoded latent feature vectors, which remain constant in time. The second group of inputs has six physics-informed temporal information arrays, which are:
\begin{enumerate}
    \item Lattice shell thickness\footnote{All scalars are expressed in the form of a constant array with a length of 50, same as the number of time steps.}.
    \item Final nominal compression strain$^1$.
    \item $\rm{log}_{\rm 10} \Dot{\bar{\epsilon}} \;^1$. The logarithm of strain rate is used instead of the strain rate since the strain rate dependence in Johnson-Cook model is logarithmic.
    \item Nominal compression strain value at each output time point.
    \item Current time value at each output time point.
    \item A binary elastic wave indicator is defined as:
    \begin{equation}
        I_e = 
        \begin{cases}
        0 ,& \quad{}  t \leq t_e\\
        1 ,& \quad{}  t > t_e\\
        \end{cases},
    \end{equation}
where $t_e = \frac{H}{\sqrt{E/\rho}}$ is an estimated time for the elastic stress wave to travel through the height of the lattice $H$. This indicates that the reaction force at the rigid bottom plate should remain 0 before the impact stress wave arrives.
\end{enumerate}
Both groups of inputs were normalized by a standard scaler in Scikit-Learn \citep{scikit-learn} prior to training. The scaler was fitted to the training data points to avoid information leakage \citep{yang2020prediction}.

To increase the amount of training data, data augmentation was applied. For each FE simulation (compressed to a constant 20\% strain), twelve final nominal strains in the range [5\%,20\%] were randomly sampled, and all inputs and outputs were linearly interpolated up to the selected final strain level. This effectively generated data points corresponding to the same strain rate but different final strain level and increased the total number of input data points to 180000. These data points were divided into training (600 designs keys, approximately 61.8\%), validation (10.9\%), testing on lattice designs seen in the training set (denoted Test1, 18.2\%), and testing on lattice designs unseen by the training set (denoted Test2, 60 design keys, approximately 9.1\%).

We employed the mean absolute error (MAE) \citep{willmott2005advantages} as the loss function, defined as:
\begin{equation}
    {\rm{MAE}} = \frac{ \sum^N_{i=1} |\bf{Y}_i - \hat{\bf{Y}}_i| }{ N },
\end{equation}
where $N,\bf{Y}_i,\hat{\bf{Y}}_i$ denote the number of training data points, ground-truth outputs, and the NN predictions, respectively. The MSE was chosen as a metric.

\subsubsection{Neural network model}
\label{sec:nn_model}
In solid mechanics, a recurrent NN model known as the gated recurrent unit (GRU) model has been widely used to predict sequences \citep{abueidda2021deep,frankel2019predicting}. As investigated by Abueidda et al. \citep{abueidda2021deep}, although the GRU model is more computationally expensive than the long short-term memory (LSTM) model and the temporal convolutional network (TCN) model, it gives most accurate predictions on energy absorption curves of an elastoplastic material undergoing complex deformations. Due to the capability of the GRU model to learn from complex deformation histories, we utilize it to predict the energy absorption of lattices under dynamic compression.

The developed GRU model consists of three stacked layers, each of 300 GRU units with hyperbolic tangent (tanh) activation, leading to a model with 1.45 million trainable parameters. The loss function was minimized using an Adam optimizer \citep{kingma2014adam} with an inverse time decay learning rate schedule and an initial learning rate of $1\times10^{-3}$. The model was trained for 150 epochs with a batch size of 600, and training was repeated 10 times to obtain average training time and model accuracy. The complete data set was reshuffled and partitioned in each training repetition, as described in \sref{sec:data_loss}. The GRU model was implemented and tested in Keras \citep{chollet2015keras} with a TensorFlow \citep{tensorflow2015-whitepaper} backend. All training was conducted on an Intel i7-11800H processor with GPU acceleration on a Nvidia GeForce RTX 3050 GPU.

\section{Results}
\label{sec:results}
\subsection{Lattice design encoding and reconstruction}
\label{sec:reconstruction}
\begin{figure}[h!]
\begin{center}
    \includegraphics[trim={3cm 5cm 4cm 5cm},clip,width=\textwidth]{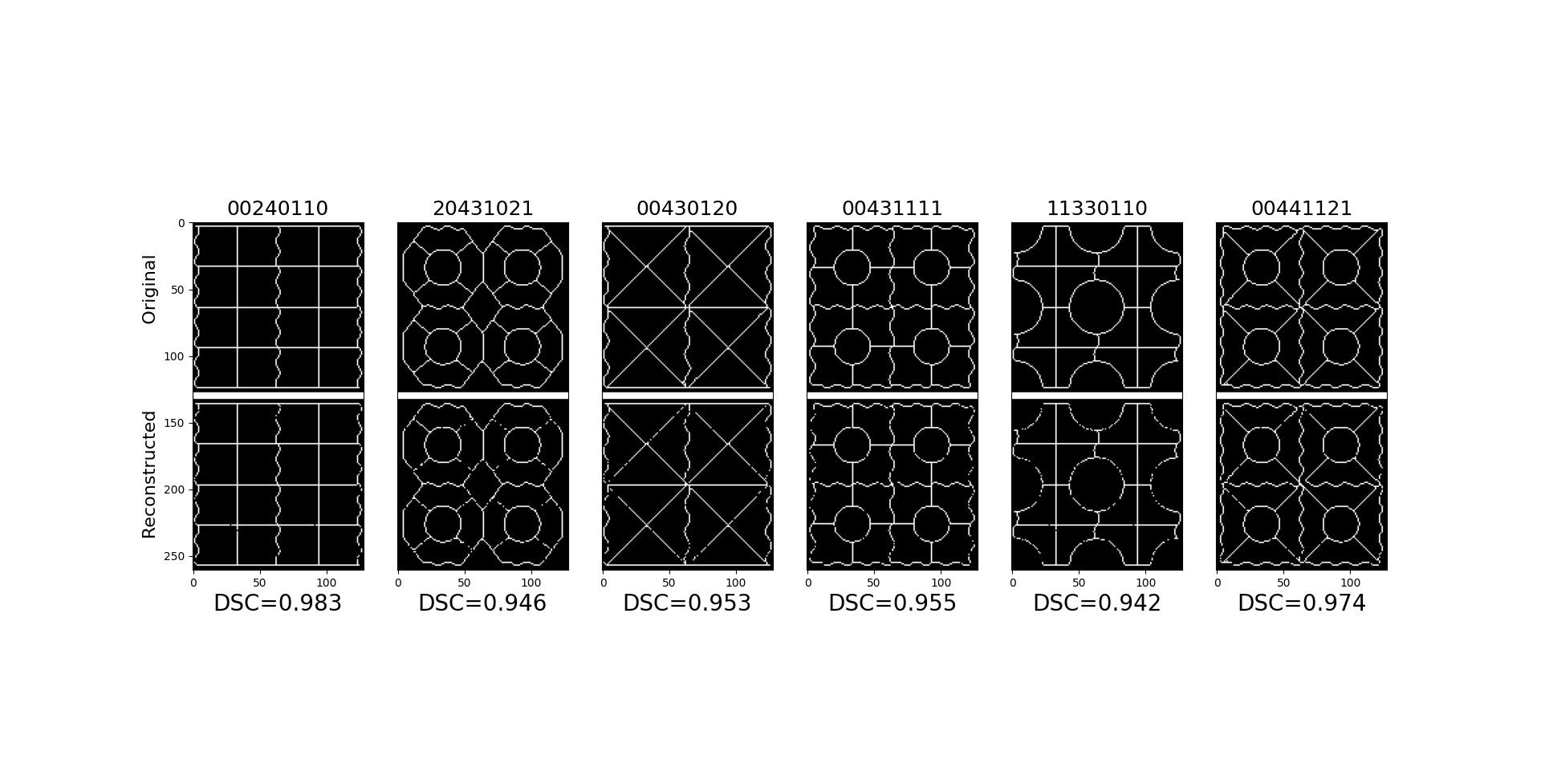} 
    \caption{Comparison of original and reconstructed lattice design skeletons for six randomly chosen designs.}
    \label{dsc}
\end{center}
\end{figure}

Using a latent space dimension of 100, we compare the original lattice designs to the reconstructed ones generated by the trained autoencoder for six randomly selected designs, see \fref{dsc}. To quantitatively describe the reconstruction quality, the dice similarity coefficient (DSC) \citep{abueidda2020topology} is used:
\begin{equation}
    {\rm{DSC}} = \frac{ 2 | \bm{I} \cap \bm{I}_r|}{ |\bm{I}| + |\bm{I}_r|},
\end{equation} 
where $\bm{I}$ and $\bm{I}_r$ are the ground-truth and reconstructed binary skeleton images (1 denotes presence of material and 0 indicates void), respectively.

\subsection{Predicting energy outputs}
\label{sec:energy}
\begin{figure}[h!] 
    \newcommand\x{0.5}
    \centering
     \subfloat[]{
         \includegraphics[trim={0 0 0cm 1cm},clip,width=\x\textwidth]{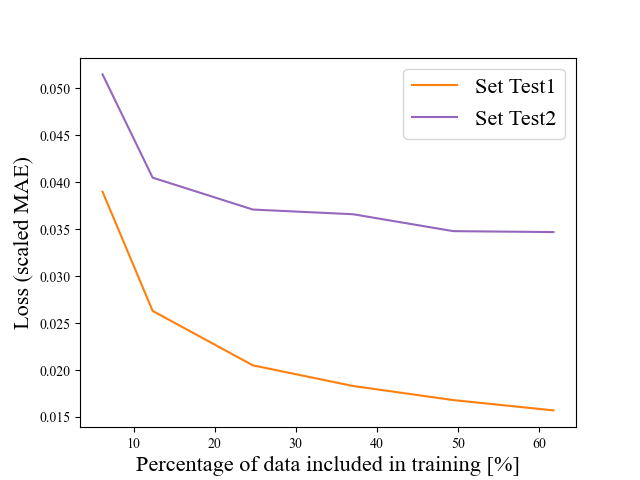}
         \label{fig:pct_plot}
     }
     \subfloat[]{
         \includegraphics[trim={0 0 0cm 1cm},clip,width=\x\textwidth]{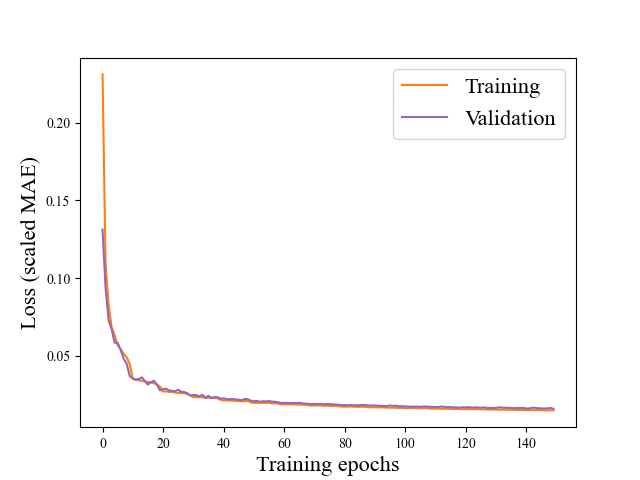}
         \label{fig:loss}
     }
    \caption{Convergence plot for GRU model training process: \psubref{fig:pct_plot} Scaled mean squared error when different percentage of the total data is used in training. \psubref{fig:loss} Scaled mean absolute error evolution during training. Note that the MAE shown here is the MAE computed on the variables scaled by the standard scaler.}
    \label{conv}
\end{figure}

To access the number of input data points required in training to obtain accurate results, different percentages of the total input data were used to train the model independently and were tested on two identical testing sets; the results are shown in \fref{fig:pct_plot}. To obtain a measure of the average performance of the GRU model, training was repeated ten times using the partition described in \sref{sec:data_loss}. A typical convergence curve showing the loss and metric during training is shown in \fref{fig:loss}.
As described in \sref{sec:data_loss}, we test our model on two testing sets: (1) Test1, which was chosen from the 600 lattice designs that were seen by training, and (2) Test2, which contains 60 lattice designs that were completely unseen by training. Mean and standard deviation of training time, relative MAE for each component in each test set, are reported in \tref{nn_stats}. Relative MAE is presented here since the four output arrays considered have vastly different scales; thus, we normalize the MAE by the range over time steps of the respective ground truth \footnote{To avoid erroneous relative MAE calculations when the range of the ground truth is very close to 0 (i.e., for cases where the stress wave has not arrived yet), a minimum range of 0.25 N is enforced for reaction force, and $1\times10^{-2}$ J for energies. }. 

\begin{table}[h]
    \caption{Mean and standard deviation of training time and relative MAE}
    \small
    \centering
    \begin{tabular}{ccccccc}
    \hline
    & \vline & \begin{tabular}{@{}c@{}}Train time\\$[$s$]$\end{tabular} & \begin{tabular}{@{}c@{}}rMAE,\\RF$^1$, $[$\%$]$\end{tabular} &
    \begin{tabular}{@{}c@{}}rMAE,\\PD$^2$, $[$\%$]$\end{tabular} &
    \begin{tabular}{@{}c@{}}rMAE,\\DMD$^3$, $[$\%$]$\end{tabular} &
    \begin{tabular}{@{}c@{}}rMAE,\\ELSE$^4$, $[$\%$]$\end{tabular} \\ 
    \hline
    Test1 & \vline & \multirow{2}{*}{3597.44 (23.15)} & 6.26 (0.19) & 0.30 (0.01) & 4.15 (0.16) & 0.95 (0.01) \\
    Test2 & \vline &  & 9.32 (1.08) & 1.04 (0.18) & 6.93 (0.61) & 2.38 (0.26) \\
    \begin{tabular}{@{}c@{}}Percent\\difference\end{tabular}  & \vline & \textbackslash & 48.75\% & 242.58\% & 67.09\% & 151.83\% \\
    \hline
    \multicolumn{7}{c}{$^1$: Reaction force; $^2$: Plastic dissipation; $^3$: Damage dissipation; $^4$: Elastic strain energy.}
    \end{tabular}
    \label{nn_stats}
\end{table}

\begin{figure}[h!]
\begin{center}
    \includegraphics[trim={4cm 1.5cm 4.5cm 2.3cm},clip,width=\textwidth]{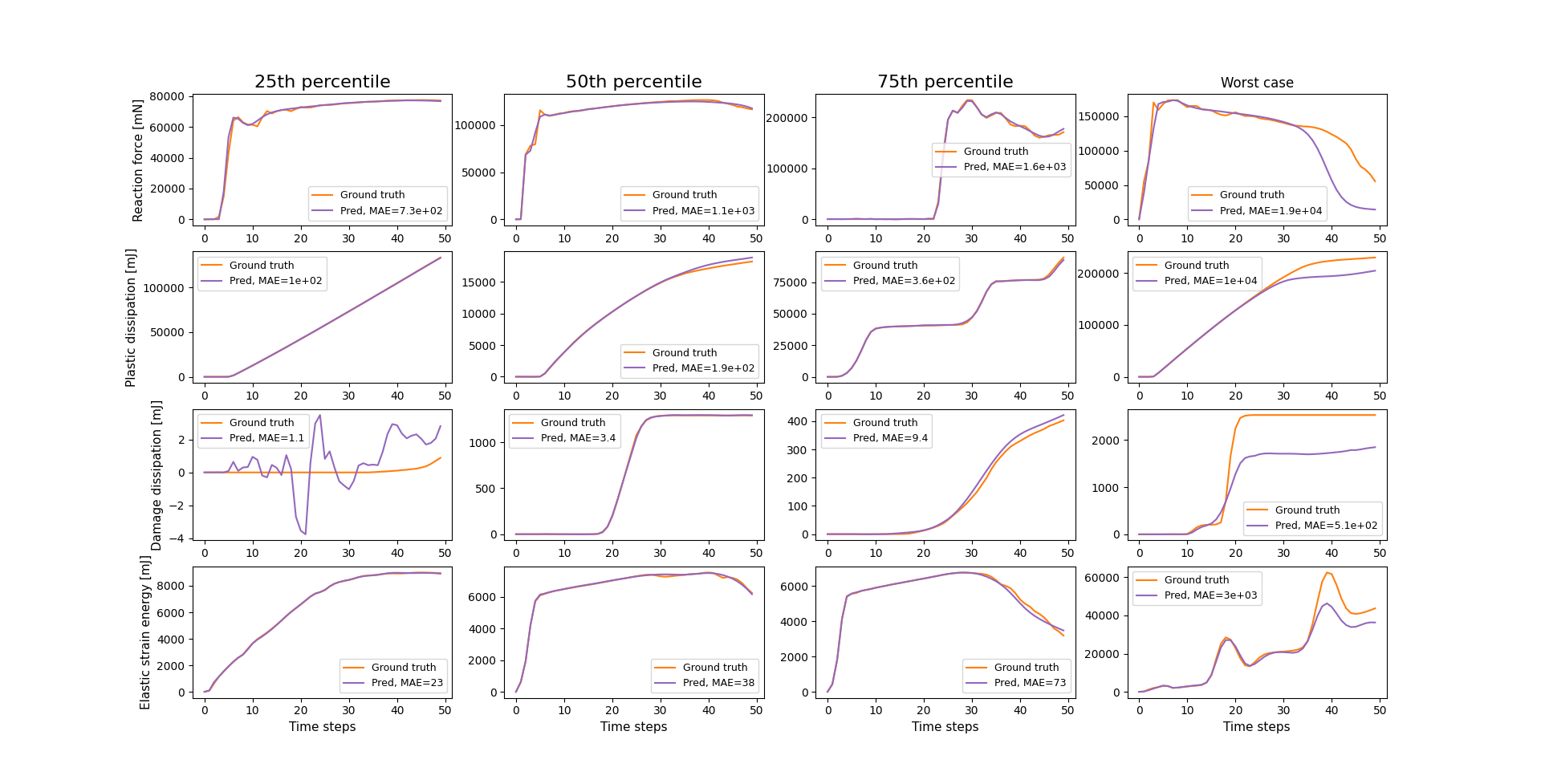} 
    \caption{Comparison of ground truths and GRU predictions on set Test1 (contains designs seen in training), ranked by percentile of MAE to provide a representative sampling. Note that the MAE is ranked independently for each of the four output arrays.}
    \label{mae_val1}
\end{center}
\end{figure}

\begin{figure}[h!]
\begin{center}
    \includegraphics[trim={4cm 1.5cm 4.5cm 2.3cm},clip,width=\textwidth]{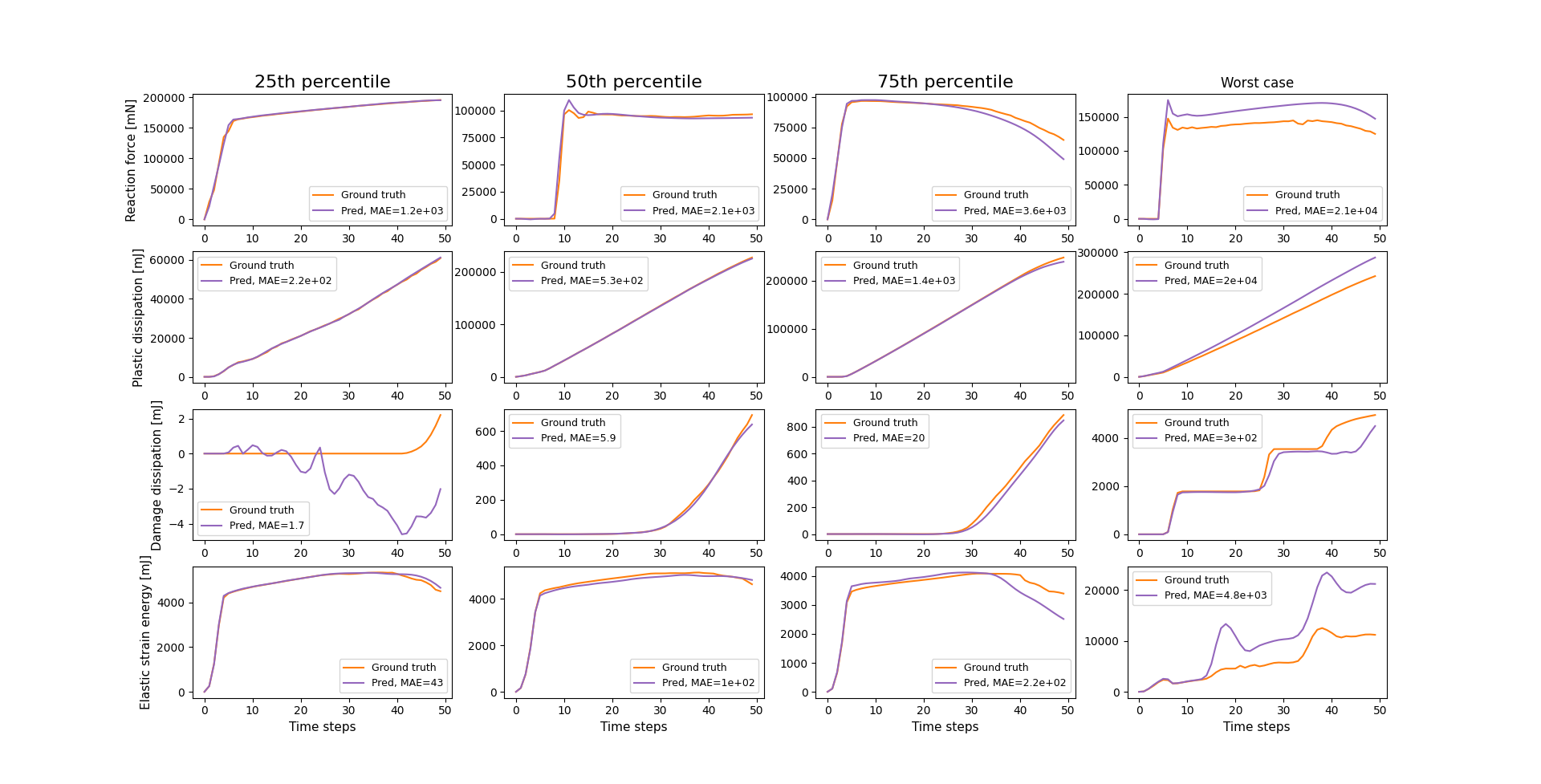} 
    \caption{Comparison of ground truths and GRU predictions on set Test2 (contains designs unseen by training), ranked by percentile of MAE to provide a representative sampling. Note that the MAE is ranked independently for each of the four output arrays.}
    \label{mae_val2}
\end{center}
\end{figure}

A comparison between ground truths and NN predictions, ranked by the percentile of MAE for each output array, is presented in Figures \ref{mae_val1} and \ref{mae_val2}. The median model (one that gives the median overall MAE among the 10 training repetitions) was used to generate the plots. 


\subsection{Extending to new lattice designs through transfer learning}
\label{sec:transfer}
Comparison of model performance between sets Test1 and Test2 shows how well the trained GRU model can generalize to previously unseen but similar designs (all were generated by the key-based system detailed in \sref{sec:geometric}). It is of interest to further investigate the performance of the as-trained model on lattice designs that are generated outside of the key-based design system. 

\begin{figure}[h!]
\begin{center}
    \includegraphics[trim={1cm 1cm 1cm 1cm},clip,width=0.7\textwidth]{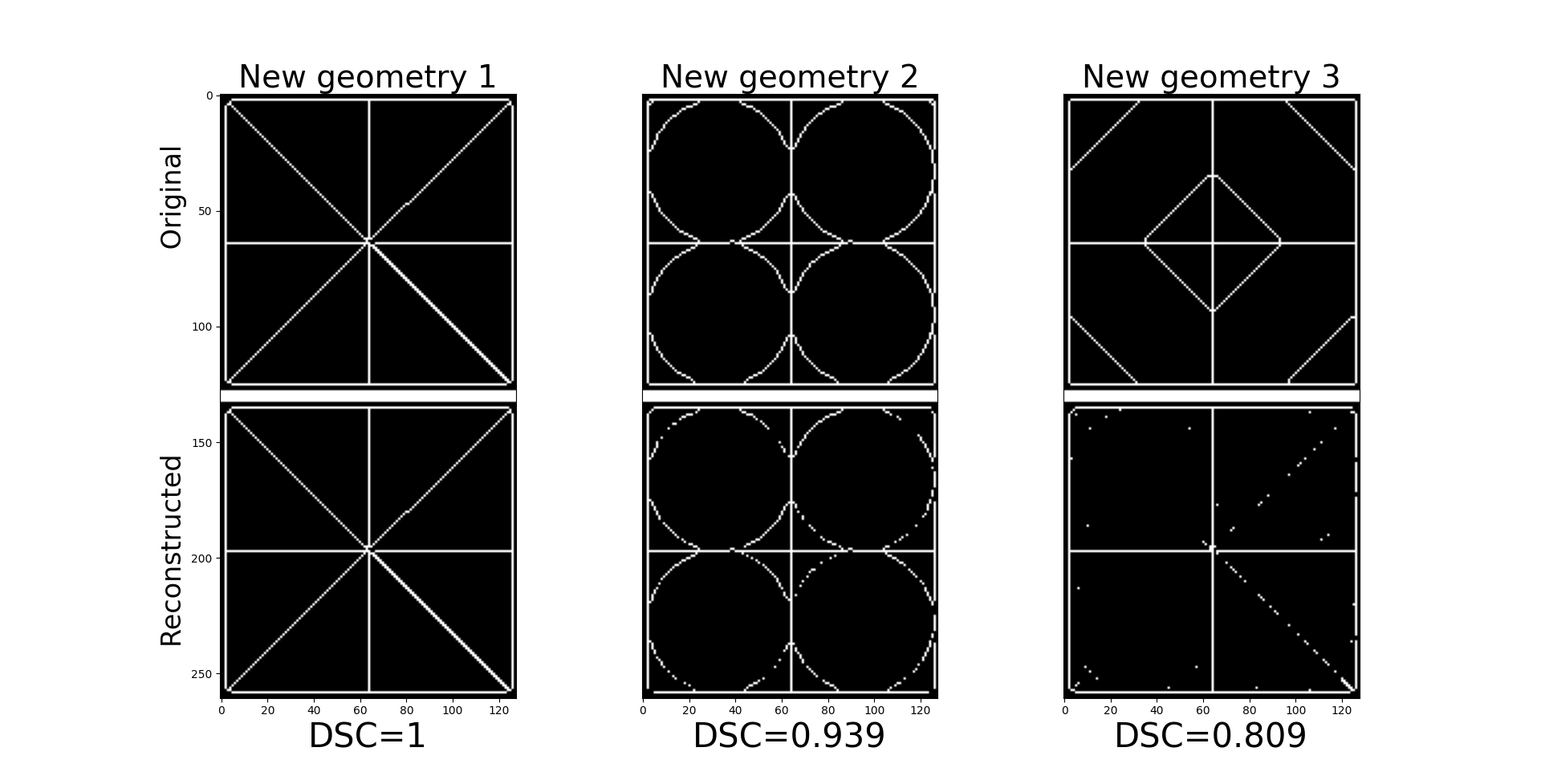} 
    \caption{Comparison of original and reconstructed lattice design skeletons for new lattice geometries.}
    \label{dsc2}
\end{center}
\end{figure}

Three new geometries were generated manually, and the comparison of the lattice design with its autoencoder reconstruction is shown in \fref{dsc2}. The mean and standard deviation of relative MAE for each output array in each test set using the as-trained model are listed in \tref{nn_stats2}. For easier comparison, results from Test2 in \tref{nn_stats} are repeated here.

\begin{table}[h]
    \caption{Mean and standard deviation of relative MAE on new geometries}
    \small
    \centering
    \begin{tabular}{cccccc}
    \hline
    & \vline & \begin{tabular}{@{}c@{}}rMAE,\\RF$^1$, $[$\%$]$\end{tabular} &
    \begin{tabular}{@{}c@{}}rMAE,\\PD$^2$, $[$\%$]$\end{tabular} &
    \begin{tabular}{@{}c@{}}rMAE,\\DMD$^3$, $[$\%$]$\end{tabular} &
    \begin{tabular}{@{}c@{}}rMAE,\\ELSE$^4$, $[$\%$]$\end{tabular} \\ 
    \hline
    Test2 & \vline & 9.32 (1.08) & 1.04 (0.18) & 6.93 (0.61) & 2.38 (0.26) \\
    
    New geometry 1 & \vline  & 132.45 (70.19) & 7.05 (0.86) & 27.59 (5.38) & 13.47 (1.03) \\
    
    New geometry 2 & \vline  & 338.09 (85.60) & 26.35 (2.05) & 38.95 (7.86) & 35.82 (1.98) \\
    
    New geometry 3 & \vline  & 186.81 (63.80) & 17.88 (1.62) & 28.52 (4.99) & 26.00 (1.30) \\
    \hline
    \multicolumn{6}{c}{$^1$: Reaction force; $^2$: Plastic dissipation; $^3$: Damage dissipation; $^4$: Elastic strain energy.}
    \end{tabular}
    \label{nn_stats2}
\end{table}

\begin{figure}[h!]
\begin{center}
    \includegraphics[trim={4cm 1.2cm 4.5cm 2.3cm},clip,width=\textwidth]{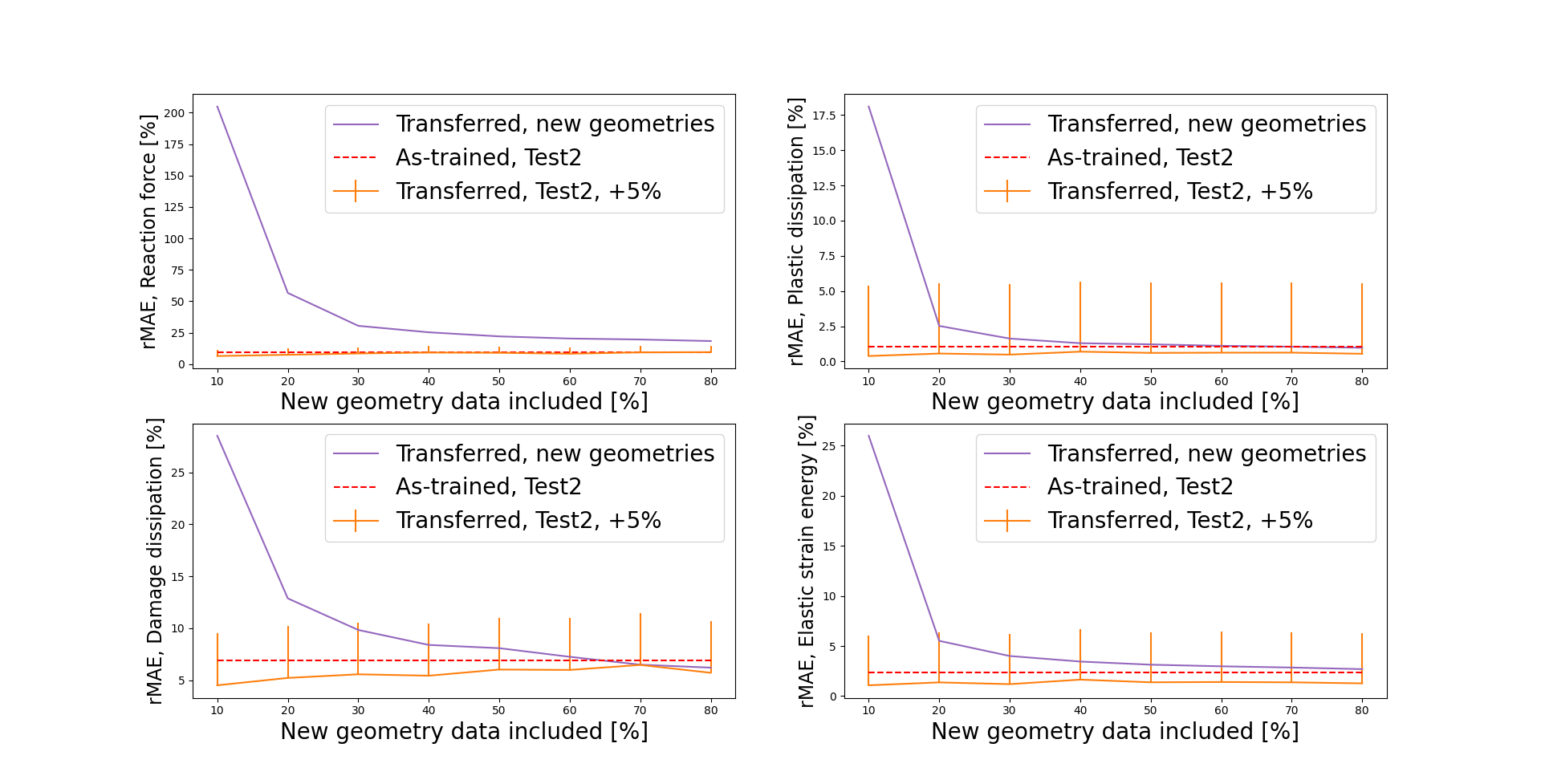} 
    \caption{Comparison of relative MAE on different output arrays using different percentages of the total data points corresponding to the new geometries. The relative MAE of the as-trained GRU model is plotted in red as a reference.}
    \label{transfer}
\end{center}
\end{figure}

\begin{figure}[h!]
\begin{center}
    \includegraphics[trim={4cm 1.5cm 4.5cm 2.3cm},clip,width=\textwidth]{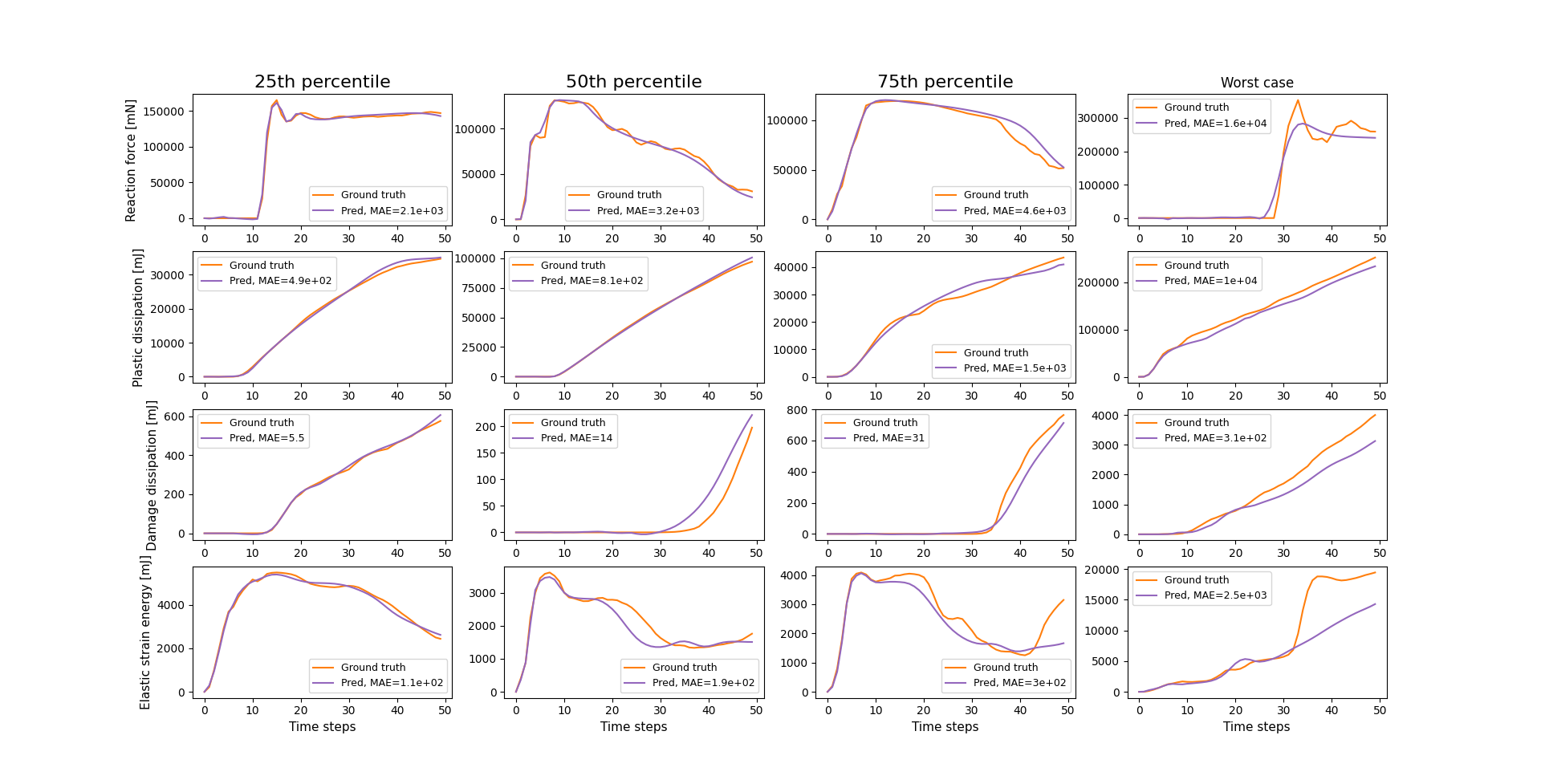} 
    \caption{Comparison of ground truths and GRU predictions on new geometries, ranked by percentile of MAE to provide a representative sampling. Note that the MAE is ranked independently for each of the four output arrays.}
    \label{transfer2}
\end{center}
\end{figure}

Besides the as-trained model, it is worth investigating how the model can be extended to unseen geometries by exposing it to a small amount of new input data via transfer learning. Transfer learning can be used to extend the trained model to a new field by transferring its knowledge learned previously from a similar field \citep{weiss2016survey}. The trained GRU model was retrained for 20 epochs on a new training set of data points corresponding to new geometries. To prevent catastrophic forgetting during transfer learning, we add 5000 original training data points corresponding to geometries in the key-based design system to the training set, although other more involved methods exist \citep{lee2017overcoming,chen2019catastrophic}. To gauge the presence and amount of catastrophic forgetting, we test our data on two testing sets: one consists of 60 lattice design keys unseen during the training of both the original GRU model and the transferred model, while the other one contains data points for the three new geometries (that are unseen by the transfer training). Fifty strain rates and thicknesses were randomly sampled from the same ranges as described in \sref{sec:fe_sim} for each of the three new geometries, leading to 150 FE simulations. A similar data augmentation technique detailed in \sref{sec:data_loss} was used to generate 7500 input data points. Plots comparing the relative MAE in both testing sets are shown in \fref{transfer}. 
A comparison between ground truths and transferred GRU model predictions, ranked by the percentile of MAE for each output array, is presented in \fref{transfer2}. The model retrained with 40\% of the new data points was used to generate the plot. 

\section{Discussion}
\label{sec:discussion}
\fref{dsc} demonstrates the ground truth images and their autoencoder reconstructions. For six randomly chosen cases, we observe a close resemblance of the two, which indicates that a 100-by-1 latent vector is sufficient to encode the input images from within the key-based design system. This confirms autoencoder as an effective tool to encode image inputs \citep{KIM2021109544,JUNG2020100690}. However, as \fref{dsc2} shows, the reconstruction quality drops significantly on images from outside of the key-based design system. Especially for the third new geometries shown in \fref{dsc2}, its autoencoder reconstruction resembles the first geometry, although the ground truths are vastly different. This confusion in the autoencoder indicates that it is specialized in encoding images within the key-based system, which all bear some design feature resemblance due to the combinatoric nature of the framework. One possible way to enhance the encoding capability of the autoencoder is to expose it to a more diverse design space. As training for the autoencoder is separated from the training of the GRU model, this additional training can be done efficiently without the need for any expensive FE simulations. It is also hypothesized that the prediction accuracy of the subsequent GRU model will increase if the autoencoder effectively and accurately extracts important latent features from the given design.

Two validation sets, Test1 and Test2, were always employed for examining the model's ability to extend the structure-property relations learned from a limited input space to new design geometries. \fref{fig:pct_plot} shows the accuracy of the two sets when a different number of data points were used in training. Although loss keeps decreasing on set Test1, model performance plateaus on set Test2 after about 50\% of points are used. For best performance in both sets, all subsequent models in this work were trained using 61.8\% of total data points, as described in \sref{sec:data_loss}. It can be inferred from \fref{fig:loss} that no severe overfitting occurred during training. Comparing the results in \tref{nn_stats}, we see that the prediction accuracy on set Test2 has deteriorated by about 50 to 240 percent compared to set Test1, but we note that all relative MAEs in Test2 remain smaller than 10\% of the ground truth range. The statistical distributions of MAEs are shown in \fref{mae_val1} and \fref{mae_val2}. Comparing the first three columns (up to 75th percentile), we observe that the GRU predictions closely follow the FE simulation results, except when the ground truth values are very small (e.g., row 3 column 1 in \fref{mae_val1}). Deviations from ground truth grow in later stages of loading, after about 30 time steps. However, even on the highest MAE case, the GRU model can predict response curves that bear similar trends as the ground truth in both testing sets. It is worth noting that a positivity constraint was not enforced on the energy outputs; thus, some non-physical negative energy values arise when the ground truths are small. In most cases, the model can accurately predict the reaction curve and energy absorption history for various lattice designs, with an observed accuracy similar to that reported by Yang et al. \citep{yang2020prediction}, which employs a convolutional NN architecture as opposed to the recurrent NN architecture in this work. As is apparent in \tref{nn_stats2}, the trained GRU model has a certain generalization ability on set Test2, but falls short on new geometries outside of the key-based design system. It is hypothesized that this limited generalization ability is due to both geometries in sets Test1 and Test2 coming from the same key-based design system and can be remedied by having a more diverse training set that covers a wider range of input designs. Previous studies also suggest alternative ways to improve the generalization ability of trained models, such as altering the input format \citep{2019arXiv190107761Z} or using a neural network ensemble \citep{6469240}.

It is unlikely that the model needs to predict mechanical response and energy absorption of a new lattice geometry given only its cross-section. It is often possible to obtain additional input data from limited testing. The model would then extend predictions to the new lattice design with different wall thicknesses and/or loading strain rates. \fref{transfer} demonstrates the results of transfer learning with different percentages of new input data. We see that transfer learning based on the GRU model trained in \sref{sec:energy} is highly effective. With as few as 20 training epochs using only 30\% of the 150 FE simulations on the new lattice designs, the prediction accuracy on the energy absorption curves is within 5\% of the model performance on the set Test2. Relative MAE of the reaction force prediction remains relatively high at 30.5\% with inclusion of 30\% new data points, but it is still a dramatic decrease from over 300\% in the as-trained model. Also, note the transferred model retains similar accuracy on set Test2 to the original GRU model, indicating that catastrophic forgetting has not occurred. \fref{transfer2} depicts the MAE distribution of the transferred model. Predictions up to the 75th percentile show relatively close agreement with FE simulation results, and large deviations only emerge after about halfway through the loading. This shows that transfer learning effectively extends a trained GRU model to new lattice designs outside of the original, finite training design space with limited new data.

\section{Conclusions and future work}
\label{sec:conc}
In this work, a GRU model was trained to predict the reaction force curve and energy absorption curves of thin-walled lattices during dynamic longitudinal compression. A novel combinatorial, key-based system was developed to generate lattice geometries that share some of the design cues with bio-inspired honeycomb lattices, which the authors believe is the first time that a combinatorial framework is used to systematically generate thin-walled lattice designs. An autoencoder was used to effectively encode the lattice cross-sectional image to reduce input dimensions. The GRU model is able to accurately predict the energy responses in unseen geometries from the key-based design system but falls short on directly extending the predicting capability to lattice designs outside of the key-based design system. Nonetheless, when enhanced with data points on the new geometries via transfer learning, the transferred GRU model reclaims much of the accuracy on new designs, while maintaining high accuracy over the designs in the key-based system. 

We conclude that the trained GRU model can accurately approximate the relations between the lattice structure and key performance measurements such as stress-strain curve and energy absorption as a function of compression strain for a wide variety of thin-walled lattices under dynamic longitudinal compression. The ability of the trained model to quickly generate performance predictions for new lattice designs even on low-end laptop platforms renders itself a suitable guide in preliminary design stages to quickly sift out potentially performant designs for more detailed analyses. 

In future work, we will focus on leveraging the gradients of the trained GRU model to inversely generate new designs with optimized specific energy absorption during high strain rate loading, similar to the work by Chen et al. \citep{chen2020generative} and Zheng et al. \citep{ZHENG2021113894}. Different ways to parameterize the lattice design space, such as using the adjacency matrix \citep{li2022encoding}, will be explored to enhance the generalizability and generative ability of future NN models. The effect of a number of unit cells and unit cell arrangements, as well as the role of periodic boundary conditions, will also be explored.

\section*{Data availability}
The data and source code that support the findings of this study can be found at: \url{https://github.com/Jasiuk-Research-Group/LatticeResponse_NN_Prediction}.

\section*{Conflict of interest}
The authors declare that they have no conflict of interest.

\section*{Acknowledgements}
We (I. J.) acknowledges the support of the Army Research Office contract (No. W 911NF-18-2-0067) and the National Science Foundation grant (MOMS-1926353).

\section*{CRediT author contributions}
\textbf{Junyan He}: Conceptualization, Methodology, Software, Formal analysis, Investigation, Data Curation, Writing - Original Draft. \textbf{Shashank Kushwaha}: Methodology, Software, Formal analysis, Investigation, Writing - Original Draft. \textbf{Diab Abueidda}: Supervision, Writing - Review \& Editing. \textbf{Iwona Jasiuk}: Supervision, Resources, Writing - Review \& Editing, Funding Acquisition.

\bibliographystyle{unsrtnat}
\setlength{\bibsep}{0.0pt}
{\scriptsize \bibliography{References.bib} }

\begin{thebibliography}{60}
\providecommand{\natexlab}[1]{#1}
\providecommand{\url}[1]{\texttt{#1}}
\expandafter\ifx\csname urlstyle\endcsname\relax
  \providecommand{\doi}[1]{doi: #1}\else
  \providecommand{\doi}{doi: \begingroup \urlstyle{rm}\Url}\fi

\bibitem[Shin et~al.(2008)Shin, Lee, and Cho]{shin2008experimental}
Kwang~Bok Shin, Jae~Youl Lee, and Se~Hyun Cho.
\newblock An experimental study of low-velocity impact responses of sandwich
  panels for korean low floor bus.
\newblock \emph{Composite Structures}, 84\penalty0 (3):\penalty0 228--240,
  2008.

\bibitem[Xie and Zhou(2014)]{xie2014impact}
Suchao Xie and Hui Zhou.
\newblock Impact characteristics of a composite energy absorbing bearing
  structure for railway vehicles.
\newblock \emph{Composites Part B: Engineering}, 67:\penalty0 455--463, 2014.

\bibitem[Van~Paepegem et~al.(2014)Van~Paepegem, Palanivelu, Degrieck, Vantomme,
  Reymen, Kakogiannis, Van~Hemelrijck, and Wastiels]{van2014blast}
Wim Van~Paepegem, Sivakumar Palanivelu, Joris Degrieck, John Vantomme, Bruno
  Reymen, Dimitrios Kakogiannis, Danny Van~Hemelrijck, and Jan Wastiels.
\newblock Blast performance of a sacrificial cladding with composite tubes for
  protection of civil engineering structures.
\newblock \emph{Composites Part B: Engineering}, 65:\penalty0 131--146, 2014.

\bibitem[Codina et~al.(2017)Codina, Ambrosini, and
  de~Borb{\'o}n]{codina2017new}
Ram{\'o}n Codina, Daniel Ambrosini, and Fernanda de~Borb{\'o}n.
\newblock New sacrificial cladding system for the reduction of blast damage in
  reinforced concrete structures.
\newblock \emph{International Journal of Protective Structures}, 8\penalty0
  (2):\penalty0 221--236, 2017.

\bibitem[Qi et~al.(2013)Qi, Yang, Yang, Wei, and Lu]{qi2013blast}
Chang Qi, Shu Yang, Li-Jun Yang, Zhi-Yong Wei, and Zhen-Hua Lu.
\newblock Blast resistance and multi-objective optimization of aluminum
  foam-cored sandwich panels.
\newblock \emph{Composite Structures}, 105:\penalty0 45--57, 2013.

\bibitem[Gama et~al.(2001)Gama, Bogetti, Fink, Yu, Claar, Eifert, and
  Gillespie~Jr]{gama2001aluminum}
Bazle~A Gama, Travis~A Bogetti, Bruce~K Fink, Chin-Jye Yu, T~Dennis Claar,
  Harald~H Eifert, and John~W Gillespie~Jr.
\newblock Aluminum foam integral armor: a new dimension in armor design.
\newblock \emph{Composite structures}, 52\penalty0 (3-4):\penalty0 381--395,
  2001.

\bibitem[Tarlochan et~al.(2007)Tarlochan, Hamouda, Mahdi, and
  Sahari]{tarlochan2007composite}
F~Tarlochan, AMS Hamouda, E~Mahdi, and BB~Sahari.
\newblock Composite sandwich structures for crashworthiness applications.
\newblock \emph{Proceedings of the Institution of Mechanical Engineers, Part L:
  Journal of Materials: Design and Applications}, 221\penalty0 (2):\penalty0
  121--130, 2007.

\bibitem[Tarlochan(2021)]{tarlochan2021sandwich}
Faris Tarlochan.
\newblock Sandwich structures for energy absorption applications: A review.
\newblock \emph{Materials}, 14\penalty0 (16):\penalty0 4731, 2021.

\bibitem[Xue and Hutchinson(2006)]{xue2006crush}
Zhenyu Xue and John~W Hutchinson.
\newblock Crush dynamics of square honeycomb sandwich cores.
\newblock \emph{International Journal for Numerical Methods in Engineering},
  65\penalty0 (13):\penalty0 2221--2245, 2006.

\bibitem[Ha et~al.(2019)Ha, Lu, and Xiang]{ha2019energy}
Ngoc~San Ha, Guoxing Lu, and Xinmei Xiang.
\newblock Energy absorption of a bio-inspired honeycomb sandwich panel.
\newblock \emph{Journal of Materials Science}, 54\penalty0 (8):\penalty0
  6286--6300, 2019.

\bibitem[Qiao and Chen(2016)]{qiao2016plane}
Jinxiu Qiao and Changqing Chen.
\newblock In-plane crushing of a hierarchical honeycomb.
\newblock \emph{International Journal of Solids and Structures}, 85:\penalty0
  57--66, 2016.

\bibitem[Paz et~al.(2014)Paz, D{\'\i}az, Romera, and Costas]{paz2014crushing}
J~Paz, J~D{\'\i}az, L~Romera, and M~Costas.
\newblock Crushing analysis and multi-objective crashworthiness optimization of
  gfrp honeycomb-filled energy absorption devices.
\newblock \emph{Finite Elements in Analysis and Design}, 91:\penalty0 30--39,
  2014.

\bibitem[Sun et~al.(2010)Sun, Li, Stone, and Li]{sun2010two}
Guangyong Sun, Guangyao Li, Michael Stone, and Qing Li.
\newblock A two-stage multi-fidelity optimization procedure for honeycomb-type
  cellular materials.
\newblock \emph{Computational Materials Science}, 49\penalty0 (3):\penalty0
  500--511, 2010.

\bibitem[Panda et~al.(2018)Panda, Leite, Biswal, Niu, and
  Garg]{panda2018experimental}
Biranchi Panda, Marco Leite, Bibhuti~Bhusan Biswal, Xiaodong Niu, and Akhil
  Garg.
\newblock Experimental and numerical modelling of mechanical properties of 3d
  printed honeycomb structures.
\newblock \emph{Measurement}, 116:\penalty0 495--506, 2018.

\bibitem[Christensen(2000)]{CHRISTENSEN200093}
R.M. Christensen.
\newblock Mechanics of cellular and other low-density materials.
\newblock \emph{International Journal of Solids and Structures}, 37\penalty0
  (1):\penalty0 93--104, 2000.
\newblock ISSN 0020-7683.
\newblock \doi{https://doi.org/10.1016/S0020-7683(99)00080-3}.
\newblock URL
  \url{https://www.sciencedirect.com/science/article/pii/S0020768399000803}.

\bibitem[Gao et~al.(2018)Gao, Li, Gao, and Xiao]{gao2018topological}
Jie Gao, Hao Li, Liang Gao, and Mi~Xiao.
\newblock Topological shape optimization of 3d micro-structured materials using
  energy-based homogenization method.
\newblock \emph{Advances in Engineering Software}, 116:\penalty0 89--102, 2018.

\bibitem[Abueidda et~al.(2020)Abueidda, Koric, and Sobh]{abueidda2020topology}
Diab~W Abueidda, Seid Koric, and Nahil~A Sobh.
\newblock Topology optimization of 2d structures with nonlinearities using deep
  learning.
\newblock \emph{Computers \& Structures}, 237:\penalty0 106283, 2020.

\bibitem[Duddeck et~al.(2016)Duddeck, Hunkeler, Lozano, Wehrle, and
  Zeng]{duddeck2016topology}
Fabian Duddeck, Stephan Hunkeler, Pablo Lozano, Erich Wehrle, and Duo Zeng.
\newblock Topology optimization for crashworthiness of thin-walled structures
  under axial impact using hybrid cellular automata.
\newblock \emph{Structural and Multidisciplinary Optimization}, 54\penalty0
  (3):\penalty0 415--428, 2016.

\bibitem[Zeng and Duddeck(2017)]{zeng2017improved}
Duo Zeng and Fabian Duddeck.
\newblock Improved hybrid cellular automata for crashworthiness optimization of
  thin-walled structures.
\newblock \emph{Structural and Multidisciplinary Optimization}, 56\penalty0
  (1):\penalty0 101--115, 2017.

\bibitem[Guo et~al.(2021)Guo, Zhao, Su, and Zuo]{guo2021topology}
Guikai Guo, Yanfang Zhao, Weihe Su, and Wenjie Zuo.
\newblock Topology optimization of thin-walled cross section using moving
  morphable components approach.
\newblock \emph{Structural and Multidisciplinary Optimization}, 63\penalty0
  (5):\penalty0 2159--2176, 2021.

\bibitem[Sharafi et~al.(2014)Sharafi, Teh, and Hadi]{sharafi2014shape}
Pezhman Sharafi, Lip~H Teh, and Muhammad~NS Hadi.
\newblock Shape optimization of thin-walled steel sections using graph theory
  and aco algorithm.
\newblock \emph{Journal of Constructional Steel Research}, 101:\penalty0
  331--341, 2014.

\bibitem[Verma et~al.(2020)Verma, Rankouhi, and Suresh]{verma2020combinatorial}
Chaman~Singh Verma, Behzad Rankouhi, and Krishnan Suresh.
\newblock A combinatorial approach for constructing lattice structures.
\newblock \emph{Journal of Mechanical Design}, 142\penalty0 (4), 2020.

\bibitem[Bastek et~al.(2022)Bastek, Kumar, Telgen, Glaesener, and
  Kochmann]{bastek2022inverting}
Jan-Hendrik Bastek, Siddhant Kumar, Bastian Telgen, Rapha{\"e}l~N Glaesener,
  and Dennis~M Kochmann.
\newblock Inverting the structure--property map of truss metamaterials by deep
  learning.
\newblock \emph{Proceedings of the National Academy of Sciences}, 119\penalty0
  (1), 2022.

\bibitem[Baykaso{\u{g}}lu et~al.(2020)Baykaso{\u{g}}lu, Baykaso{\u{g}}lu, and
  Cetin]{baykasouglu2020multi}
Adil Baykaso{\u{g}}lu, Cengiz Baykaso{\u{g}}lu, and Erhan Cetin.
\newblock Multi-objective crashworthiness optimization of lattice structure
  filled thin-walled tubes.
\newblock \emph{Thin-Walled Structures}, 149:\penalty0 106630, 2020.

\bibitem[Wang et~al.(2021)Wang, Callanan, Ogunbodede, and
  Rai]{wang2021hierarchical}
Jun Wang, Jesse Callanan, Oladapo Ogunbodede, and Rahul Rai.
\newblock Hierarchical combinatorial design and optimization of non-periodic
  metamaterial structures.
\newblock \emph{Additive Manufacturing}, 37:\penalty0 101710, 2021.

\bibitem[Callanan et~al.(2018)Callanan, Ogunbodede, Dhameliya, Wang, and
  Rai]{callanan2018hierarchical}
Jesse Callanan, Oladapo Ogunbodede, Maulikkumar Dhameliya, Jun Wang, and Rahul
  Rai.
\newblock Hierarchical combinatorial design and optimization of quasi-periodic
  metamaterial structures.
\newblock In \emph{International Design Engineering Technical Conferences and
  Computers and Information in Engineering Conference}, volume 51760, page
  V02BT03A011. American Society of Mechanical Engineers, 2018.

\bibitem[Wang et~al.(2018)Wang, Zhou, Khaliulin, and Shabalov]{wang2018six}
ZhiJin Wang, HuaZhi Zhou, VI~Khaliulin, and AV~Shabalov.
\newblock Six-ray folded configurations as the geometric basis of thin-walled
  elements in engineering structures.
\newblock \emph{Thin-Walled Structures}, 130:\penalty0 435--448, 2018.

\bibitem[Yang et~al.(2020)Yang, Kim, Ryu, and Gu]{yang2020prediction}
Charles Yang, Youngsoo Kim, Seunghwa Ryu, and Grace~X Gu.
\newblock Prediction of composite microstructure stress-strain curves using
  convolutional neural networks.
\newblock \emph{Materials \& Design}, 189:\penalty0 108509, 2020.

\bibitem[Gu et~al.(2018)Gu, Chen, and Buehler]{gu2018novo}
Grace~X Gu, Chun-Teh Chen, and Markus~J Buehler.
\newblock De novo composite design based on machine learning algorithm.
\newblock \emph{Extreme Mechanics Letters}, 18:\penalty0 19--28, 2018.

\bibitem[Chen and Gu(2019)]{chen2019machine}
Chun-Teh Chen and Grace~X Gu.
\newblock Machine learning for composite materials.
\newblock \emph{MRS Communications}, 9\penalty0 (2):\penalty0 556--566, 2019.

\bibitem[Yang et~al.(2019)Yang, Kim, Ryu, and Gu]{yang2019using}
Charles Yang, Youngsoo Kim, Seunghwa Ryu, and Grace~X Gu.
\newblock Using convolutional neural networks to predict composite properties
  beyond the elastic limit.
\newblock \emph{MRS Communications}, 9\penalty0 (2):\penalty0 609--617, 2019.

\bibitem[Abueidda et~al.(2019)Abueidda, Almasri, Ammourah, Ravaioli, Jasiuk,
  and Sobh]{ABUEIDDA2019111264}
Diab~W. Abueidda, Mohammad Almasri, Rami Ammourah, Umberto Ravaioli, Iwona~M.
  Jasiuk, and Nahil~A. Sobh.
\newblock Prediction and optimization of mechanical properties of composites
  using convolutional neural networks.
\newblock \emph{Composite Structures}, 227:\penalty0 111264, 2019.
\newblock ISSN 0263-8223.
\newblock \doi{https://doi.org/10.1016/j.compstruct.2019.111264}.
\newblock URL
  \url{https://www.sciencedirect.com/science/article/pii/S0263822319312383}.

\bibitem[Laban et~al.(2020)Laban, Gowid, and Mahdi]{laban2020experimental}
Othman Laban, Samer Gowid, and Elsadig Mahdi.
\newblock Experimental investigation and uncertainty prediction of the
  load-carrying capacity of composite double hat for lattice core sandwich
  panels using artificial neural network.
\newblock In \emph{2020 IEEE International Conference on Informatics, IoT, and
  Enabling Technologies (ICIoT)}, pages 67--72. IEEE, 2020.

\bibitem[Messner(2019)]{10.1115/1.4045040}
Mark~C. Messner.
\newblock {Convolutional Neural Network Surrogate Models for the Mechanical
  Properties of Periodic Structures}.
\newblock \emph{Journal of Mechanical Design}, 142\penalty0 (2), 10 2019.
\newblock ISSN 1050-0472.
\newblock \doi{10.1115/1.4045040}.
\newblock URL \url{https://doi.org/10.1115/1.4045040}.
\newblock 024503.

\bibitem[Garland et~al.(2020)Garland, White, Jared, Heiden, Donahue, and
  Boyce]{garland2020deep}
Anthony~P Garland, Benjamin~C White, Bradley~H Jared, Michael Heiden, Emily
  Donahue, and Brad~L Boyce.
\newblock Deep convolutional neural networks as a rapid screening tool for
  complex additively manufactured structures.
\newblock \emph{Additive Manufacturing}, 35:\penalty0 101217, 2020.

\bibitem[Hassanin et~al.(2020)Hassanin, Alkendi, Elsayed, Essa, and
  Zweiri]{hassanin2020controlling}
Hany Hassanin, Yusra Alkendi, Mahmoud Elsayed, Khamis Essa, and Yahya Zweiri.
\newblock Controlling the properties of additively manufactured cellular
  structures using machine learning approaches.
\newblock \emph{Advanced Engineering Materials}, 22\penalty0 (3):\penalty0
  1901338, 2020.

\bibitem[Zok et~al.(2016)Zok, Latture, and Begley]{zok2016periodic}
Frank~W Zok, Ryan~M Latture, and Matthew~R Begley.
\newblock Periodic truss structures.
\newblock \emph{Journal of the Mechanics and Physics of Solids}, 96:\penalty0
  184--203, 2016.

\bibitem[Yang et~al.(2018)Yang, Sun, Yang, and Pan]{yang2018out}
Xianfeng Yang, Yuxin Sun, Jialing Yang, and Qifan Pan.
\newblock Out-of-plane crashworthiness analysis of bio-inspired aluminum
  honeycomb patterned with horseshoe mesostructure.
\newblock \emph{Thin-Walled Structures}, 125:\penalty0 1--11, 2018.

\bibitem[San~Ha and Lu(2020)]{san2020review}
Ngoc San~Ha and Guoxing Lu.
\newblock A review of recent research on bio-inspired structures and materials
  for energy absorption applications.
\newblock \emph{Composites Part B: Engineering}, 181:\penalty0 107496, 2020.

\bibitem[SIMULIA(2021)]{Abaqus2021}
SIMULIA.
\newblock Abaqus, 2021.

\bibitem[Johnson and Cook(1985)]{johnson1985fracture}
Gordon~R Johnson and William~H Cook.
\newblock Fracture characteristics of three metals subjected to various
  strains, strain rates, temperatures and pressures.
\newblock \emph{Engineering Fracture Mechanics}, 21\penalty0 (1):\penalty0
  31--48, 1985.

\bibitem[Wang and Shi(2013)]{wang2013validation}
Xuemei Wang and Jun Shi.
\newblock Validation of johnson-cook plasticity and damage model using impact
  experiment.
\newblock \emph{International Journal of Impact Engineering}, 60:\penalty0
  67--75, 2013.

\bibitem[Othman and Marzouk(2016)]{othman2016strain}
Hesham Othman and H~Marzouk.
\newblock Strain rate sensitivity of fiber-reinforced cementitious composites.
\newblock \emph{American Concrete Institute Materials Journal}, 113\penalty0
  (2):\penalty0 143--150, 2016.

\bibitem[Chollet et~al.(2015)]{chollet2015keras}
Francois Chollet et~al.
\newblock Keras, 2015.
\newblock URL \url{https://github.com/fchollet/keras}.

\bibitem[Abadi et~al.(2015)Abadi, Agarwal, Barham, Brevdo, Chen, Citro,
  Corrado, Davis, Dean, Devin, Ghemawat, Goodfellow, Harp, Irving, Isard, Jia,
  Jozefowicz, Kaiser, Kudlur, Levenberg, Man\'{e}, Monga, Moore, Murray, Olah,
  Schuster, Shlens, Steiner, Sutskever, Talwar, Tucker, Vanhoucke, Vasudevan,
  Vi\'{e}gas, Vinyals, Warden, Wattenberg, Wicke, Yu, and
  Zheng]{tensorflow2015-whitepaper}
Mart\'{i}n Abadi, Ashish Agarwal, Paul Barham, Eugene Brevdo, Zhifeng Chen,
  Craig Citro, Greg~S. Corrado, Andy Davis, Jeffrey Dean, Matthieu Devin,
  Sanjay Ghemawat, Ian Goodfellow, Andrew Harp, Geoffrey Irving, Michael Isard,
  Yangqing Jia, Rafal Jozefowicz, Lukasz Kaiser, Manjunath Kudlur, Josh
  Levenberg, Dandelion Man\'{e}, Rajat Monga, Sherry Moore, Derek Murray, Chris
  Olah, Mike Schuster, Jonathon Shlens, Benoit Steiner, Ilya Sutskever, Kunal
  Talwar, Paul Tucker, Vincent Vanhoucke, Vijay Vasudevan, Fernanda Vi\'{e}gas,
  Oriol Vinyals, Pete Warden, Martin Wattenberg, Martin Wicke, Yuan Yu, and
  Xiaoqiang Zheng.
\newblock {TensorFlow}: Large-scale machine learning on heterogeneous systems,
  2015.
\newblock URL \url{https://www.tensorflow.org/}.
\newblock Software available from tensorflow.org.

\bibitem[Kingma and Ba(2014)]{kingma2014adam}
Diederik~P Kingma and Jimmy Ba.
\newblock Adam: A method for stochastic optimization.
\newblock \emph{arXiv preprint arXiv:1412.6980}, 2014.

\bibitem[Pedregosa et~al.(2011)Pedregosa, Varoquaux, Gramfort, Michel, Thirion,
  Grisel, Blondel, Prettenhofer, Weiss, Dubourg, Vanderplas, Passos,
  Cournapeau, Brucher, Perrot, and Duchesnay]{scikit-learn}
F.~Pedregosa, G.~Varoquaux, A.~Gramfort, V.~Michel, B.~Thirion, O.~Grisel,
  M.~Blondel, P.~Prettenhofer, R.~Weiss, V.~Dubourg, J.~Vanderplas, A.~Passos,
  D.~Cournapeau, M.~Brucher, M.~Perrot, and E.~Duchesnay.
\newblock Scikit-learn: Machine learning in {P}ython.
\newblock \emph{Journal of Machine Learning Research}, 12:\penalty0 2825--2830,
  2011.

\bibitem[Willmott and Matsuura(2005)]{willmott2005advantages}
Cort~J Willmott and Kenji Matsuura.
\newblock Advantages of the mean absolute error (mae) over the root mean square
  error (rmse) in assessing average model performance.
\newblock \emph{Climate research}, 30\penalty0 (1):\penalty0 79--82, 2005.

\bibitem[Abueidda et~al.(2021)Abueidda, Koric, Sobh, and
  Sehitoglu]{abueidda2021deep}
Diab~W Abueidda, Seid Koric, Nahil~A Sobh, and Huseyin Sehitoglu.
\newblock Deep learning for plasticity and thermo-viscoplasticity.
\newblock \emph{International Journal of Plasticity}, 136:\penalty0 102852,
  2021.

\bibitem[Frankel et~al.(2019)Frankel, Jones, Alleman, and
  Templeton]{frankel2019predicting}
Ari~L Frankel, Reese~E Jones, Coleman Alleman, and Jeremy~A Templeton.
\newblock Predicting the mechanical response of oligocrystals with deep
  learning.
\newblock \emph{Computational Materials Science}, 169:\penalty0 109099, 2019.

\bibitem[Weiss et~al.(2016)Weiss, Khoshgoftaar, and Wang]{weiss2016survey}
Karl Weiss, Taghi~M Khoshgoftaar, and DingDing Wang.
\newblock A survey of transfer learning.
\newblock \emph{Journal of Big Data}, 3\penalty0 (1):\penalty0 1--40, 2016.

\bibitem[Lee et~al.(2017)Lee, Kim, Jun, Ha, and Zhang]{lee2017overcoming}
Sang-Woo Lee, Jin-Hwa Kim, Jaehyun Jun, Jung-Woo Ha, and Byoung-Tak Zhang.
\newblock Overcoming catastrophic forgetting by incremental moment matching.
\newblock \emph{Advances in neural information processing systems}, 30, 2017.

\bibitem[Chen et~al.(2019)Chen, Wang, Fu, Long, and Wang]{chen2019catastrophic}
Xinyang Chen, Sinan Wang, Bo~Fu, Mingsheng Long, and Jianmin Wang.
\newblock Catastrophic forgetting meets negative transfer: Batch spectral
  shrinkage for safe transfer learning.
\newblock \emph{Advances in Neural Information Processing Systems}, 32, 2019.

\bibitem[Kim et~al.(2021)Kim, Park, Jung, Asghari-Rad, Lee, Kim, Jung, and
  Kim]{KIM2021109544}
Yongju Kim, Hyung~Keun Park, Jaimyun Jung, Peyman Asghari-Rad, Seungchul Lee,
  Jin~You Kim, Hwan~Gyo Jung, and Hyoung~Seop Kim.
\newblock Exploration of optimal microstructure and mechanical properties in
  continuous microstructure space using a variational autoencoder.
\newblock \emph{Materials \& Design}, 202:\penalty0 109544, 2021.
\newblock ISSN 0264-1275.
\newblock \doi{https://doi.org/10.1016/j.matdes.2021.109544}.
\newblock URL
  \url{https://www.sciencedirect.com/science/article/pii/S0264127521000976}.

\bibitem[Jung et~al.(2020)Jung, Yoon, Park, Jo, and Kim]{JUNG2020100690}
Jaimyun Jung, Jae~Ik Yoon, Hyung~Keun Park, Hyeontae Jo, and Hyoung~Seop Kim.
\newblock Microstructure design using machine learning generated low
  dimensional and continuous design space.
\newblock \emph{Materialia}, 11:\penalty0 100690, 2020.
\newblock ISSN 2589-1529.
\newblock \doi{https://doi.org/10.1016/j.mtla.2020.100690}.
\newblock URL
  \url{https://www.sciencedirect.com/science/article/pii/S2589152920301071}.

\bibitem[{Zhang} et~al.(2019){Zhang}, {Peng}, {Zhou}, {Xiang}, and
  {Wang}]{2019arXiv190107761Z}
Yiquan {Zhang}, Bo~{Peng}, Xiaoyi {Zhou}, Cheng {Xiang}, and Dalei {Wang}.
\newblock {A deep Convolutional Neural Network for topology optimization with
  strong generalization ability}.
\newblock \emph{arXiv e-prints}, art. arXiv:1901.07761, January 2019.

\bibitem[Yang et~al.(2013)Yang, Zeng, Zhong, and Wu]{6469240}
Jing Yang, Xiaoqin Zeng, Shuiming Zhong, and Shengli Wu.
\newblock Effective neural network ensemble approach for improving
  generalization performance.
\newblock \emph{IEEE Transactions on Neural Networks and Learning Systems},
  24\penalty0 (6):\penalty0 878--887, 2013.
\newblock \doi{10.1109/TNNLS.2013.2246578}.

\bibitem[Chen and Gu(2020)]{chen2020generative}
Chun-Teh Chen and Grace~X Gu.
\newblock Generative deep neural networks for inverse materials design using
  backpropagation and active learning.
\newblock \emph{Advanced Science}, 7\penalty0 (5):\penalty0 1902607, 2020.

\bibitem[Zheng et~al.(2021)Zheng, Kumar, and Kochmann]{ZHENG2021113894}
Li~Zheng, Siddhant Kumar, and Dennis~M. Kochmann.
\newblock Data-driven topology optimization of spinodoid metamaterials with
  seamlessly tunable anisotropy.
\newblock \emph{Computer Methods in Applied Mechanics and Engineering},
  383:\penalty0 113894, 2021.
\newblock ISSN 0045-7825.
\newblock \doi{https://doi.org/10.1016/j.cma.2021.113894}.
\newblock URL
  \url{https://www.sciencedirect.com/science/article/pii/S0045782521002310}.

\bibitem[Li et~al.(2022)Li, Liu, Chen, Jiang, Nie, and Pan]{li2022encoding}
Shunning Li, Yuanji Liu, Dong Chen, Yi~Jiang, Zhiwei Nie, and Feng Pan.
\newblock Encoding the atomic structure for machine learning in materials
  science.
\newblock \emph{Wiley Interdisciplinary Reviews: Computational Molecular
  Science}, 12\penalty0 (1):\penalty0 e1558, 2022.

\end{thebibliography}
\end{document}